\documentclass{article}
\usepackage{epsfig,rotating,amsmath,alltt,boxedminipage,times,url}
\usepackage{natbib}
\bibpunct[; ]{(}{)}{,}{a}{}{;}

\title{A Formal Framework for Linguistic Annotation}
\author{Steven Bird and Mark Liberman\\
  Linguistic Data Consortium, University of Pennsylvania\\
  3615 Market Street, Philadelphia, PA 19104-2608, USA
}
\date{May 2, 2000}

\newenvironment{sv}{\scriptsize\begin{alltt}}{\end{alltt}\normalsize}

\usepackage{fancyheadings}
\def\myurl#1{{\small\url{#1}}}

\def\glb{\operatorname{glb}}
\def\lub{\operatorname{lub}}

\def\au#1#2{\ar@/^/[#1]^{#2}}
\def\ad#1#2{\ar@/_/[#1]_{#2}}
\def\larc#1{\ar@/^/[r]^{#1}}
\def\arc{\ar@/^/[r]}
\def\smtt#1{{\small\tt #1}}

\begin{document}
\maketitle\thispagestyle{empty}

\begin{abstract}{\normalsize
`Linguistic annotation' covers any descriptive or analytic 
notations applied to raw language data. The basic data may be in the form of 
time functions -- audio, video and/or physiological recordings -- or it may 
be textual. The added notations may include transcriptions of all sorts (from 
phonetic features to discourse structures), part-of-speech and sense tagging, 
syntactic analysis, `named entity' identification, co-reference annotation, 
and so on.
While there are several ongoing efforts to provide formats
and tools for such annotations and to publish annotated linguistic
databases, the lack of widely accepted standards is becoming a critical
problem.  Proposed standards, to the extent they exist, have focused
on file formats.  This paper focuses instead on the logical
structure of linguistic annotations.  We survey a wide variety of
existing annotation formats and demonstrate a common conceptual core,
the {\em annotation graph}.  This
provides a formal framework for constructing, maintaining and searching
linguistic annotations,
while remaining consistent with many alternative data structures
and file formats.

\centerline{\bf Zusammenfassung}

Der Begriff `Linguistische Annotation' bezeichnet alle Arten
deskriptiver oder analytischer Beschreibung von Sprachdaten. Die
Ausgangsdaten k\"onnen dabei entweder die Form von Zeitfunktionen
haben -- also z.B. Audio, Video und/oder physiologische Signale --
oder als Text vorliegen. Die Annotation dagegen kann folgende
Inhalte haben: alle Arten von Transkriptionen (von phonetischen
Merkmalen bis zu Dialog-Strukturen), Phrasen- oder
Inhalts-Segmentierung, syntaktische Analysen, Identifikation von
`named entities', Querverweise innerhalb der Annotation, usw. Zwar
stehen zur Zeit mehrere verschiedene Formate und Werkzeuge zur
linguistischen Annotation zur Verf\"ugung, andererseits entwickelt
sich das Fehlen eines allgemein akzeptierten Standards zu einem
ernsten Problem. Bisher vorgeschlagene Standards konzentrieren
sich auf die Datenformate. Dieser Beitrag dagegen konzentriert
sich auf die logische Struktur linguistischer Annotationen. Wir
untersuchen eine breite Auswahl existierender Formate und k\"onnen
zeigen, da\ss\ diesen ein gemeinsames Konzept zugrundeliegt. Dieses
bildet die Grundlage f\"ur einen algebraischen Formalismus zur
linguistischen Annotation, w\"ahrend gleichzeitig die Konsistenz zu
vielen alternativen Datenstrukturen und Datenformaten erhalten
bleibt.

\centerline{\bf R\'esum\'e}

Par $\ll$ annotation linguistique $\gg$ nous d\'esignons toute
notation descriptive ou analytique appliqu\'ee \`a des donn\'ees
langagi\`eres brutes.  Ces donn\'ees brutes peuvent \^etre des signaux
temporels -- enregistrements audio, vid\'eo et/ou physiologiques -- ou
du texte.  Les notations ajout\'ees peuvent \^etre des transcriptions
de toute nature (des traits phon\'etiques aux structures du discours),
des cat\'egories grammaticales ou s\'emantiques, une analyse
syntaxique, l'identification d'$\ll$ entit\'es nomm\'ees $\gg$,
l'annotation de co-r\'ef\'erences, etc.  Malgr\'e les efforts
entrepris pour cr\'eer des formats et des outils adapt\'es \`a de
telles annotations et pour diffuser des bases de donn\'ees
linguistiques annot\'ees, le manque de standards largement accept\'es
devient un probl\`eme critique.  Les standards propos\'es, lorsqu'ils
existent, se concentrent sur les formats de fichiers.  Cet article se
concentre au contraire sur la structure logique des annotations
linguistiques.  Nous passons en revue une grande vari\'et\'e de
formats d'annotations existants et en d\'egageons une structure
conceptuelle commune, le graphe d'annotation.  Ceci fournit un cadre
formel pour construire des annotations linguistiques, les tenir \`a
jour et y effectuer des requ\`etes, tout en restant coh\'erent avec de
nombreux autres structures de donn\'ees et formats de fichiers.
\vspace{1ex}

}
\end{abstract}

\section{Introduction}

In the simplest and commonest case, `linguistic annotation' is an
orthographic transcription of speech, time-aligned to an audio or
video recording.  Other central examples include morphological
analysis, part-of-speech tagging and syntactic bracketing; phonetic
segmentation and labeling; annotation of disfluencies, prosodic
phrasing, intonation, gesture, and discourse structure; marking of
co-reference, `named entity' tagging, and sense tagging; and
phrase-level or word-level translations.  Linguistic annotations may
describe texts or recorded signals. Our focus will be on the latter,
broadly construed to include any kind of audio, video or physiological
recording, or any combination of these, for which we will use the
cover term `linguistic signals'. However, our ideas also apply to the
annotation of texts.


Linguistic annotations have seen increasingly broad use in the scientific study
of language, in research and development of language-related
technologies, and in language-related applications more broadly, for
instance in the entertainment industry. Particular cases range from
speech databases used in speech recognition or speech synthesis
development, to annotated ethnographic materials, to cartoon sound
tracks. There have been many independent efforts to provide tools for
creating linguistic annotations, to provide general formats
for expressing them, and to provide tools for creating, browsing and
searching databases containing them -- see [\myurl{www.ldc.upenn.edu/annotation/}].
Within the area of speech and
language technology development alone, hundreds of annotated
linguistic databases have been published in the past fifteen years.


While the utility of existing tools, formats and databases is
unquestionable, their sheer variety -- and the lack of standards able
to mediate among them -- is becoming a critical problem.
Particular bodies of data are
created with particular needs in mind, using formats and tools
tailored to those needs, based on the resources and practices of the
community involved.  Once created, a linguistic database
may subsequently be used for a variety of unforeseen purposes,
both inside and outside the community that created it.  Adapting
existing software for creation, update, indexing, search and display
of `foreign' databases typically requires extensive
re-engineering. Working across a set of databases requires repeated
adaptations of this kind.

As we survey speech transcription and annotation across many
existing `communities of practice', we observe a rich diversity of
concrete format.
Various attempts to standardize practice have focused directly
on these file formats and on the tags and attributes for
describing content.
However, we contend that file formats and content
specifications are secondary.
Instead, we focus on the logical structure of linguistic
annotations, since it is here that we observe a striking commonality.
We describe a simple formal framework having a practically useful
formal structure.  This opens up an interesting range of new
possibilities for creation, maintenance and search.  We claim that
essentially all existing annotations can be expressed in this
framework. Thus, the framework should provide a useful `interlingua'
for translation among the multiplicity of current annotation formats,
and also should permit the development of new tools with broad
applicability.

This distinction between data formats and logical structure
can be brought into sharp focus by analogy with
database systems.  Consider the relationship between the abstract
notion of a relational algebra, the features of a relational database
system, and the characteristics of a particular database.  For
example, the definition of substantive notions like `date' does not
belong in the relational algebra, though there is good reason for a database
system to have a special data type for dates.  Moreover, a
particular database may incorporate all manner of restrictions on
dates and relations among them.  The formalization presented here is
targeted at the most abstract level: we want to get the annotation
formalism right.  We assume that system implementations will add all
kinds of special-case data types (i.e.\ types of labels with
specialized syntax and semantics).  We further assume that particular
databases will want to introduce additional specifications.

\begin{figure}
\centerline{
  \epsfig{figure=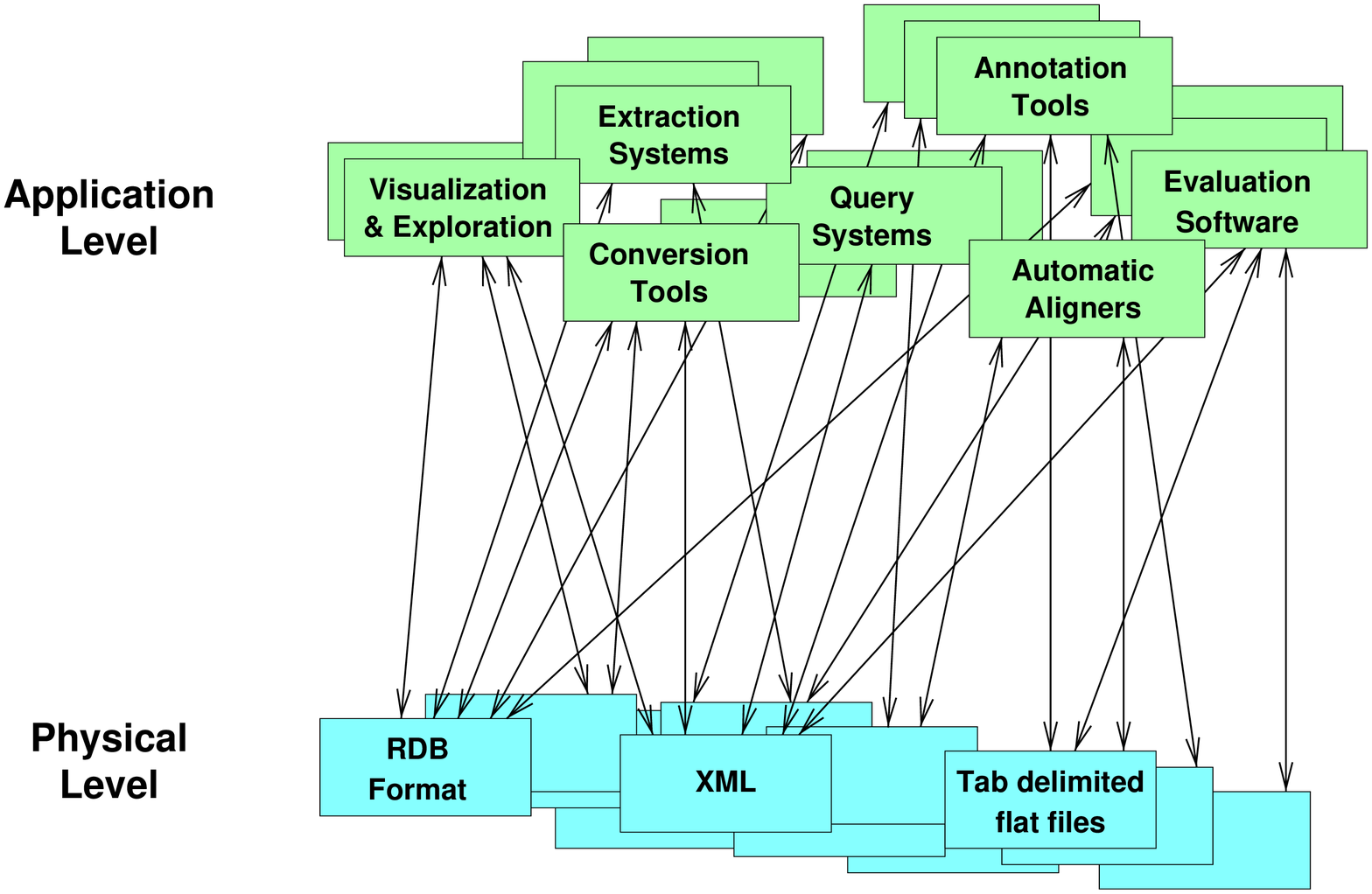,width=0.45\linewidth}
  \hfil
  \epsfig{figure=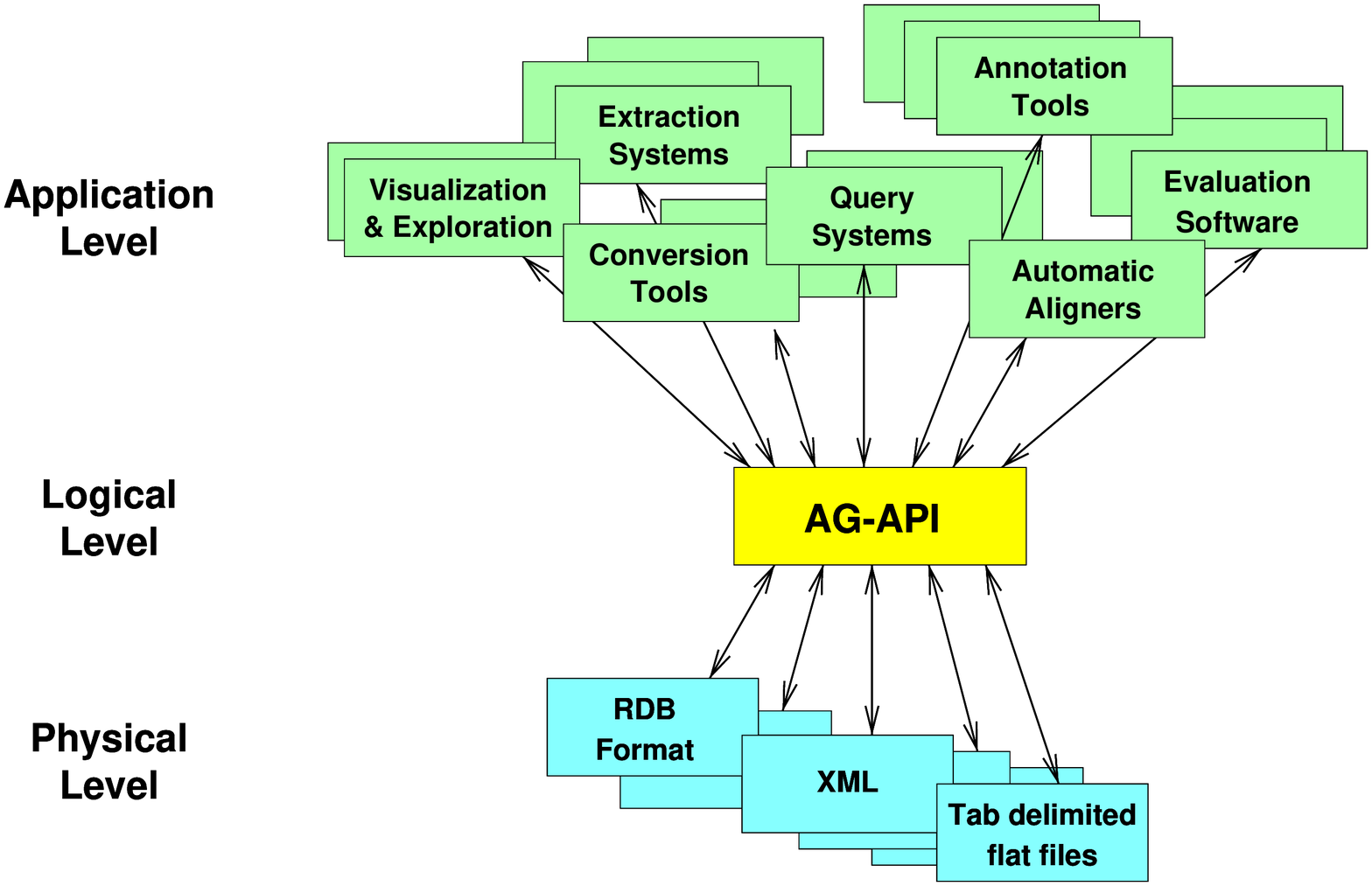,width=0.45\linewidth}
}
\caption{The Two and Three-Level Architectures for Speech Annotation}\label{fig:arch}
\vspace*{2ex}\hrule
\end{figure}

In the early days of database systems, data manipulation required
explicit reference to physical storage in files, and application
software had to be custom-built.  In the late 1960s, with the
development of the so-called ``three-level architecture'',
database functionalities were divided into three levels:
physical, logical and external.  Here, we
apply the same development to databases of annotated speech.
Figure~\ref{fig:arch} depicts the speech annotation version
of the three-level architecture.

This model permits users to create and manipulate annotation
data in the way that conforms most closely to their own conception
of the structure of the underlying data, to the contingencies of
the task at hand, and to individual preference.
Furthermore, it is possible to change an implementation at the
physical
level while leaving the higher levels intact -- i.e.\ the
{\it data independence principle}.
By adopting this model, the volatile nature of formats and the
open-ended issues associated with user interfaces no longer
present barriers on the road towards standardization.
In fact, a large number of tools will be able to comprehend a large
number of formats, so tools can interoperate and formats are
translatable.  Therefore communities wedded to particular
formats or tools are not left out in the cold.

Before we embark on our survey, a terminological aside is necessary.
As far as we are aware, there is no existing cover term for the kinds
of transcription, description and analysis that we address here.
`Transcription' may refer to the use of ordinary orthography, or a
phonetic orthography; it can plausibly be extended to certain aspects
of prosody (`intonational transcription'), but not to other kinds of
analysis (morphological, syntactic, rhetorical or discourse
structural, semantic, etc).  One does not talk about a `syntactic
transcription', although this is at least as determinate a
representation of the speech stream as is a phonetic transcription.
`Coding' has been used by social scientists to mean something like
`the assignment of events to stipulated symbolic categories,' as a
generalization of the ordinary language meaning associated with
translating words and phrases into references to a shared, secret code
book.  It would be idiosyncratic and confusing (though conceptually
plausible) to refer to ordinary orthographic transcription in this
way.  The term `markup' has come to have a specific technical meaning,
involving the addition of typographical or structural information to a
document.

In ordinary language, `annotation' means a sort of commentary or
explanation (typically indexed to particular portions of a text), or
the act of producing such a commentary.  Like `markup', this term's
ordinary meaning plausibly covers the non-transcriptional kinds of
linguistic analysis, such as the annotation of syntactic structure or
of co-reference. Some speech and language engineers have begun to use
`annotation' in this way, but there is not yet a specific,
widely-accepted technical meaning. We feel that it is reasonable to
generalize this term to cover the case of transcribing speech, by
thinking of `annotation' as the provision of any symbolic description
of particular portions of a pre-existing linguistic object.
If the object is a speech recording, then an ordinary orthographic
transcription is certainly a kind of annotation in this sense --
though it is one in which the amount of critical judgment is
small.

In sum, `annotation' is a reasonable candidate for adoption as the
needed cover term.  The alternative would be to create a neologism
(`scription'?).  Extension of the existing term `annotation' seems
preferable to us.

\section{Existing Annotation Systems}\label{sec:survey}

In order to justify our claim that essentially all existing linguistic
annotations can be expressed in the framework that we propose, we need
to discuss a representative set of such annotations. In addition,
it will be easiest to understand our proposal if we motivate it,
piece by piece, in terms of the logical structures underlying
existing annotation practice.
	
This section reviews several bodies of annotation practice, with
a concrete example of each.
For each example, we show how to express its various structuring
conventions in terms of our `annotation graphs', which are
networks consisting of nodes and arcs, decorated with
time marks and labels.  Following the
review, we shall discuss some general architectural issues (\S\ref{sec:arch}),
give a formal presentation of the `annotation graph'
concept (\S\ref{sec:algebra}).
The paper concludes in \S\ref{sec:conclusion} with an evaluation
of the formalism and a discussion of future work.

The annotation models to be discussed in detail are
TIMIT \citep{TIMIT86},
Partitur \citep{Schiel98},
CHILDES \citep{MacWhinney95},
LACITO \citep{Jacobson00},
LDC Telephone Speech,
NIST UTF \citep{UTF98},
Switchboard \citep{Godfrey92},
and MUC-7 Coreference \citep{Hirschman97}.
Three general purpose models will also be discussed
in brief: Emu \citep{CassidyHarrington00},
Festival \citep{Taylor00},
MATE \citep{McKelvie00}.
These models are widely divergent in type and purpose.
Some, like TIMIT, are associated with a specific database,
others, like UTF, are associated with a specific linguistic
domain (here conversation),
while still others, like Festival, are associated with a
specific application domain (here, speech synthesis).

Several other systems and formats have been considered in developing our
ideas, but will not be discussed in detail. These include
Switchboard \citep{Godfrey92},
HCRC MapTask \citep{Anderson91},
and TEI \citep{TEI-P3}.
The Switchboard and MapTask formats are conversational transcription
systems that encode a subset of the information in the LDC and NIST
formats cited above. The TEI guidelines for `Transcriptions of
Speech' \cite[p11]{TEI-P3} are also similar in content, though they
offer access to a very broad range of representational techniques
drawn from other aspects of the TEI specification. The TEI report
sketches or alludes to a correspondingly wide range of possible issues
in speech annotation.  All of these seem to be encompassed within our
proposed framework, but it does not seem appropriate to speculate at
much greater length about this, given that this portion of the TEI
guidelines does not seem to have been used in any published
transcriptions to date.
Many other models exist \citep{Altosaar98,Hertz90,Schegloff98}
and space limits our treatment of them here.

Note that there are many kinds of linguistic database that are not
linguistic annotations in our sense, although they may be connected
with linguistic annotations in various ways.  One example is a lexical
database with pointers to speech recordings along with transcriptions
of those recordings (e.g.\ HyperLex, \citealt{Bird97sigphon}).
Another example would be collections of
information that are not specific to any particular stretch of speech,
such as demographic information about speakers.
We return to such cases in \S\ref{sec:extensions}.

\subsection{TIMIT}\label{sec:timit}

The TIMIT corpus of read speech
was designed to provide data for the acquisition of
acoustic-phonetic knowledge and to support the development
and evaluation of automatic speech recognition systems.
TIMIT was the first annotated speech database to be widely distributed,
and it has been widely used and also republished in several
different forms.  It is also especially simple and clear
in structure.
Here, we just give one example taken from the TIMIT database
\citep{TIMIT86}.

\begin{figure}
\begin{sv}
train/dr1/fjsp0/sa1.wrd:              train/dr1/fjsp0/sa1.phn:
2360 5200 she                         0 2360 h#
5200 9680 had                         2360 3720 sh
9680 11077 your                       3720 5200 iy
11077 16626 dark                      5200 6160 hv
16626 22179 suit                      6160 8720 ae
22179 24400 in                        8720 9680 dcl
24400 30161 greasy                    9680 10173 y
30161 36150 wash                      10173 11077 axr
36720 41839 water                     11077 12019 dcl
41839 44680 all                       12019 12257 d
44680 49066 year                      ...
\end{sv}

\centerline{\epsfig{figure=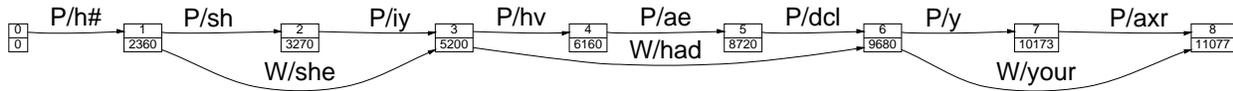,width=0.95\linewidth}}
\caption{TIMIT Annotation Data and Graph Structure}\label{timit}
\vspace*{2ex}\hrule
\end{figure}

The \smtt{.wrd} file in Figure~\ref{timit}
combines an ordinary string of orthographic words with
information about the starting and ending time of each word,
measured in audio samples at a sampling rate of 16 kHz.
The path name \smtt{train/dr1/fjsp0/sa1.wrd} tells us that
this is training data, from `dialect region 1', from female speaker
`jsp0', containing words and audio sample numbers.
The \smtt{.phn} file contains a corresponding
broad phonetic transcription.

We can interpret each line: \smtt{$<$time1$>$ $<$time2$>$ $<$label$>$}
as an edge in a directed acyclic graph, where
the two times are attributes of nodes
and the label is a property of an edge connecting those nodes.
The resulting annotation graph for the above fragment is
shown in Figure~\ref{timit}.  Observe that edge labels
have the form \smtt{$<$type$>$/$<$content$>$} where
the \smtt{$<$type$>$} here tells us what kind of label it is.
We have used \smtt{P} for the (phonetic transcription) contents of the \smtt{.phn}
file, and \smtt{W} for the (orthographic word) contents of the \smtt{.wrd} file.
The top number for each node is an identifier,
while the bottom number is the time reference.

\subsection{Partitur}\label{sec:partitur}

The Partitur format of the Bavarian Archive for Speech Signals
\citep{Schiel98} is founded on the collective experience of a broad
range of German speech database efforts.  The aim has been to create
`an open (that is extensible), robust format to represent results from
many different research labs in a common source.'  Partitur is
valuable because it represents a careful attempt to present a common
low-level core for all of those independent efforts, similar in spirit
to our effort here.  In essence, Partitur extends and reconceptualizes the TIMIT
format to encompass a wide range of annotation types.

\begin{figure}
\begin{sv}
KAN: 0 j'a:      ORT: 0 ja         TRL: 0 <A>           MAU: 4160 1119 0 j
KAN: 1 S'2:n@n   ORT: 1 sch"onen   TRL: 0 ja ,          MAU: 5280 2239 0 a:
KAN: 2 d'aNk     ORT: 2 Dank       TRL: 1 sch"onen      MAU: 7520 2399 1 S
KAN: 3 das+      ORT: 3 das        TRL: 1 <:<#Klopfen>  MAU: 9920 1599 1 2:
KAN: 4 vE:r@+    ORT: 4 w"are      TRL: 2 Dank:> ,      MAU: 11520 479 1 n
KAN: 5 z'e:6     ORT: 5 sehr       TRL: 3 das           MAU: 12000 479 1 n
KAN: 6 n'Et      ORT: 6 nett       TRL: 4 w"ar'         MAU: 12480 479 -1 <nib>
\end{sv}

\centerline{\epsfig{figure=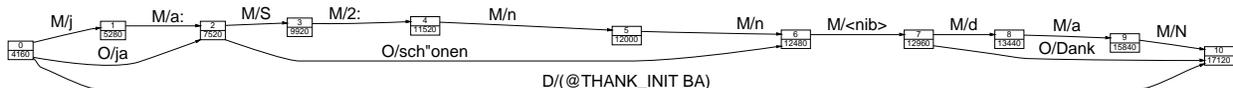,width=0.95\linewidth}}
\caption{BAS Partitur Annotation Data and Graph Structure}\label{partitur}
\vspace*{2ex}\hrule
\end{figure}

The Partitur format permits time-aligned, multi-tier description of speech
signals, along with links between units on different tiers
which are independent of the temporal structure.
For ease of presentation, the example Partitur file will be broken
into a number of chunks, and certain details (such as the header)
will be ignored.  The fragment under discussion is from one
of the Verbmobil corpora at the Bavarian Archive of Speech Signals.
The KAN tier provides the
canonical transcription, and introduces a numerical identifier
for each word to serve as an anchor for all other material.
Tiers for orthography (ORT), transliteration (TRL), and phonetic
segments (MAU) reference these anchors, using the second-last field
in each case.  The first seven lines of
information for each tier are given in Figure~\ref{partitur}.

The additional numbers for the MAU tier give offset and duration
information.  Higher level structure representing dialogue acts
refers to extended intervals using
contiguous sequences of anchors, as shown below:

\begin{sv}
DAS: 0,1,2 @(THANK_INIT BA)
DAS: 3,4,5,6 @(FEEDBACK_ACKNOWLEDGEMENT BA)
\end{sv}

The content of the first few words of the
ORT (orthography), DAS (dialog act) and MAU
(phonetic segment) tiers can
apparently be expressed as in Figure~\ref{partitur}.
Note that we abbreviate the types, using
\smtt{O/} for ORT, \smtt{D/} for DAS, and \smtt{M/} for MAU.

\subsection{CHILDES}
\label{sec:childes}

\begin{figure}
\begin{sv}
@Begin                                              *ROS:   yahoo.
@Filename:      boys73.cha                          %snd:   "boys73a.aiff" 7349 8338
@Participants:  ROS Ross Child, MAR Mark Child,     *FAT:   you got a lot more to do # don't you?
                FAT Brian Father, MOT Mary Mother   %snd:   "boys73a.aiff" 8607 9999
@Date:  4-APR-1984                                  *MAR:   yeah.
@Age of ROS:    6;3.11                              %snd:   "boys73a.aiff" 10482 10839
@Sex of ROS:    Male                                *MAR:   because I'm not ready to go to
@Birth of ROS:  25-DEC-1977                                 <the bathroom> [>] +/.
@Age of MAR:    4;4.15                              %snd:   "boys73a.aiff" 11621 13784
@Birth of MAR:  19-NOV-1979                         ...
@Sex of MAR:    male
@Situation:     Room cleaning
\end{sv}

\centerline{\epsfig{figure=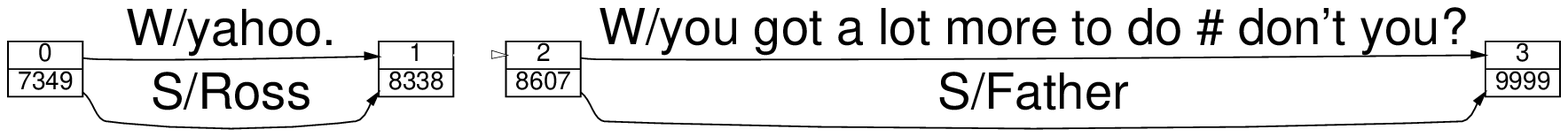,width=0.65\linewidth}}
\vspace*{2ex}
\centerline{\epsfig{figure=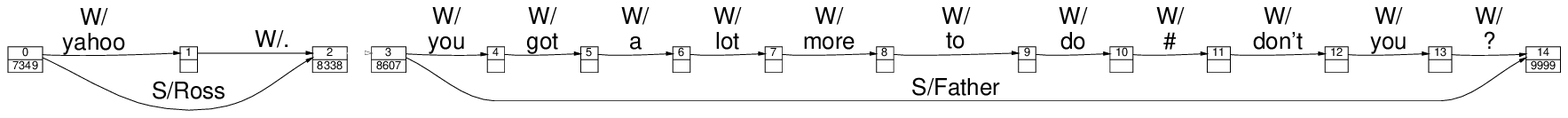,width=0.95\linewidth}}
\caption{CHILDES Annotation Data and Graph Structure}\label{chat}
\vspace*{2ex}\hrule
\end{figure}

With its extensive user base, tools and documentation,
and its coverage of some two dozen languages,
the Child Language Data Exchange System, or CHILDES,
represents the largest scientific -- as opposed to
engineering -- enterprise involved in our survey.
The CHILDES database includes a vast amount of transcript data
collected from children and adults who are learning languages
\citep{MacWhinney95}.  
All of the data are transcribed in the so-called `CHAT' format;
a typical instance is provided by
the opening fragment of a CHAT transcription shown in Figure~\ref{chat}.

The \smtt{\%snd} lines, by the conventions of this notation, provide
times for the previous transcription lines, in milliseconds relative
to the beginning of the referenced file. The first two lines of this
transcript might then be represented as the first graph in Figure~\ref{chat}.
However, this representation treats
entire phrases as atomic arc labels, complicating indexing and search.
We favor the representation in the second graph in Figure~\ref{chat},
where labels have
uniform ontological status regardless of the presence vs.\ absence of
time references.  Observe that most of the nodes in the second version
{\it could} have been given time references in the CHAT format but
were not.  The graph structure remains the same regardless of the
sparseness of temporal information.

Some of the tokens of the transcript, i.e.\ the punctuation marks,
do not reference stretches of time in the
same way that orthographic words do.  Accordingly, they may be
given a different type, and/or assigned to an instant rather than
a period (see \S\ref{sec:issues}).

\subsection{LACITO Linguistic Data Archiving Project}
\label{sec:lacito}

\begin{figure}
\begin{sv}
<HEADER> <TITLE>Deux s?x0153;urs.</TITLE> <SOUNDFILE href="SOEURS.mp2"/> </HEADER>
<BODY lang="hayu">
  <S id="s1"> <AUDIO start="2.3656" end="7.9256"/>
    <TRANSCR> <W><FORM>nakpu</FORM><GLS>deux</GLS></W>
              <W><FORM>nonotso</FORM><GLS>s?x0153;urs</GLS></W>
              <W><FORM>si?x014b;</FORM><GLS>bois</GLS></W>
              <W><FORM>pa</FORM><GLS>faire</GLS></W>
              <W><FORM>la?x0294;natshem</FORM><GLS>all\`erent(D)</GLS></W>
              <W><FORM>are</FORM><GLS>dit.on</GLS></W>
              <PONCT>.</PONCT> </TRANSCR>
    <TRADUC lang="Francais">On raconte que deux soeurs all\`erent chercher du bois.</TRADUC>
    <TRADUC lang="Anglais">They say that two sisters went to get firewood.</TRADUC>
  </S>
...
\end{sv}

\centerline{\epsfig{figure=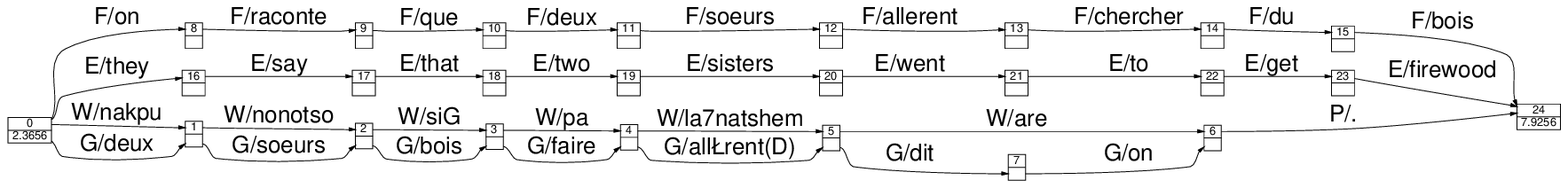,width=\linewidth}}
\caption{LACITO Annotation Data and Graph Structure}\label{archivage}
\vspace*{2ex}\hrule
\end{figure}

LACITO -- Langues et Civilisations \`a Tradition Orale -- is a CNRS
organization concerned with research on unwritten languages.  The
LACITO Linguistic Data Archiving Project was founded to conserve and
distribute the large quantity of recorded, transcribed speech data
collected by LACITO members over the last three decades
\citep{Jacobson00}.  The annotation model uses XML, and different
XSL stylesheets provide a variety of views on the base data.

In this section we discuss a transcription for
an utterance in Hayu, a Tibeto-Burman language of Nepal.  The gloss
and free translation are in French.  Consider the XML annotation
data and the graph structure in Figure~\ref{archivage}.
Here we have three types of edge labels: \smtt{W/} for the wordforms of the
Hayu story; \smtt{G/} for the gloss, and
\smtt{F/}, \smtt{E/} for phrasal translations into French and English.
In this example, the time references (which are in seconds)
are again given only
at the beginning and end of the phrase, as required by the
LACITO format.  Nevertheless, the individual Hayu words have
temporal extent and one might want to indicate that in the annotation.
Observe that there is no meaningful way of assigning time
references to word boundaries in the phrasal translation, or
for the boundary in the gloss for \smtt{dit.on}.
Thus the omission of time references may happen because
the times are simply unknown, as in Figure~\ref{chat},
or are inappropriate, as in Figure~\ref{archivage}.

\subsection{LDC Telephone Speech Transcripts}
\label{sec:callhome}

The Linguistic Data Consortium (LDC) is an open consortium of
universities, companies and government research laboratories, hosted
by the University of Pennsylvania, that creates, collects and
publishes speech and text databases, lexicons, and similar resources.
Since its foundation in 1992, it has published some 150 digital
databases, most of which contain material that falls under our
definition of `linguistic annotation.'

The LDC-published CALLHOME corpora include digital audio, 
transcripts and lexicons for telephone conversations in several languages
[\myurl{www.ldc.upenn.edu/Catalog/LDC96S46.html}].
The corpora are designed to support research on speech recognition algorithms.
The transcripts exhibit abundant overlap between speaker turns
in two-way telephone conversations.

\begin{figure}
\begin{sv}
962.68 970.21 A: He was changing projects every couple of weeks and he
  said he couldn't keep on top of it. He couldn't learn the whole new area  
968.71 969.00 B: %mm.  
970.35 971.94 A: that fast each time.  
971.23 971.42 B: %mm.    
972.46 979.47 A: %um, and he says he went in and had some tests, and he
  was diagnosed as having attention deficit disorder. Which  
980.18 989.56 A: you know, given how he's how far he's gotten, you know,
  he got his degree at &Tufts and all, I found that surprising that for
  the first time as an adult they're diagnosing this. %um  
989.42 991.86 B: %mm. I wonder about it. But anyway.  
991.75 994.65 A: yeah, but that's what he said. And %um  
994.19 994.46 B: yeah.  
995.21 996.59 A: He %um  
996.51 997.61 B: Whatever's helpful.  
997.40 1002.55 A: Right. So he found this new job as a financial
  consultant and seems to be happy with that.  
1003.14 1003.45 B: Good.  
\end{sv}
\vspace*{2ex}

\centerline{\epsfig{figure=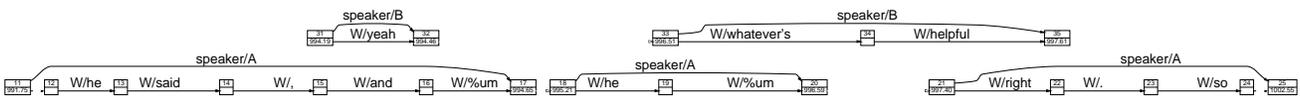,width=\linewidth}}
\caption{LDC Telephone Speech Data and Graph Structure}\label{callhome}
\vspace*{2ex}\hrule
\end{figure}

Figure~\ref{callhome} gives a typical fragment of an annotation.  Each stretch of
speech consists of a begin time, an end time, a speaker designation
(`A' or `B' in the example below), and the transcription for the
cited stretch of time.  Observe that speaker turns may be partially
or totally overlapping.

Long turns (e.g.\ the period from 972.46 to 989.56 seconds) were
broken up into shorter stretches for the convenience of the
annotators.  Thus, this format is ambiguous as to whether adjacent
stretches by the same speaker should be considered parts of the same
unit, or parts of different units.  However, the intent
is clearly just to provide additional time references within long
turns, so the most appropriate choice seems to be to merge abutting
same-speaker structures while retaining the additional time-marks.

A section of this annotation which includes an example of total
overlap is represented as an annotation graph in the lower half
of Figure~\ref{callhome}.
Turns are attributed to speakers using the \smtt{speaker/} type.
All of the words, punctuation and disfluencies are given the \smtt{W/}
type, though we could easily opt for a more refined version in which
these are assigned different types.
Observe that the annotation graph representation preserves the
non-explicitness of the original file format concerning which
of speaker A's words overlap which of speaker B's words. Of course,
additional time references could specify the overlap down to any
desired level of detail.

\subsection{NIST Universal Transcription Format}

\begin{figure}
\begin{sv}
<turn speaker="Roger_Hedgecock" spkrtype="male" dialect="native"
    startTime="2348.811875" endTime="2391.606000" mode="spontaneous" fidelity="high">
  ...
  <time sec="2378.629937">
  now all of those things are in doubt after forty years of democratic rule in
  <b_enamex type="ORGANIZATION">congress<e_enamex>
  <time sec="2382.539437">
  \{breath because <contraction e_form="[you=>you]['ve=>have]">you've got quotas
  \{breath and set<hyphen>asides and rigidities in this system that keep you
  <time sec="2387.353875">
  on welfare and away from real ownership
  \{breath and <contraction e_form="[that=>that]['s=>is]">that's a real problem in this
  <b_overlap startTime="2391.115375" endTime="2391.606000">country<e_overlap>
</turn>
<turn speaker="Gloria_Allred" spkrtype="female" dialect="native"
    startTime="2391.299625" endTime="2439.820312" mode="spontaneous" fidelity="high">
  <b_overlap startTime="2391.299625" endTime="2391.606000">well i<e_overlap>
  think the real problem is that %uh these kinds of republican attacks
  <time sec="2395.462500">
  i see as code words for discrimination
  ...
</turn>
\end{sv}

\centerline{\epsfig{figure=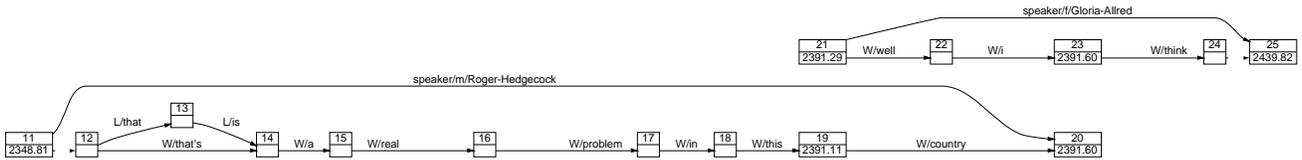,width=\linewidth}}
\caption{UTF Annotation Data and Graph Structure}\label{utf}
\vspace*{2ex}\hrule
\end{figure}

The US National Institute of Standards and Technology (NIST) has
developed a set of annotation conventions `intended to
provide an extensible universal format for transcription and
annotation across many spoken language technology evaluation domains'
\citep{UTF98}.  This `Universal Transcription Format' (UTF) was based
on the LDC Broadcast News format. A key design
goal for UTF was to provide an SGML-based format that would cover both
the LDC broadcast transcriptions and also various LDC-published
conversational transcriptions, while also providing for plausible
extensions to other sorts of material.
A notable aspect of UTF is its treatment of overlapping speaker turns.
Figure~\ref{utf} contains a fragment of UTF, taken from
the Hub-4 1997 evaluation set.

Figure~\ref{utf} contains two speaker turns, where the first speaker's
utterance of `country' overlaps the second speaker's utterance of
`well I' (note that overlaps are marked with
\smtt{$<$b\_overlap$>$} (begin overlap) and
\smtt{$<$e\_overlap$>$} (end overlap) tags)
Note that the time attributes for overlap are not required
to coincide, since they are aligned to `the most inclusive word
boundaries for each speaker turn involved in the overlap'.  The
coincidence of end times here is probably an artifact of the system
used to create the annotations.

The structure of overlapping turns can be represented using an annotation
graph as shown in Figure~\ref{utf}.  Each speaker turn is a separate
connected subgraph, disconnected from other speaker turns.
The time courses of independent utterances are logically asynchronous,
and so we prefer not to convolve them into a single stream, as the
SGML representation does.  Observe that the information
about overlap is now implicit in the time references.
Partial word overlap can also be represented if necessary.
This seems like the
best choice in general, since there is no necessary logical structure
to conversational overlaps -- at base, they are just two different
actions unfolding over the same time period.
The cited annotation graph structure is thus less explicit about word
overlaps than the UTF file.\footnote{However, if a more explicit symbolic
representation of overlaps is desired, specifying that such-and-such a
stretch of one speaker turn is associated with such-and-such a stretch
of another speaker turn, this can be represented in our framework
using the inter-arc linkage method described in \S\ref{sec:equivalence}.}

Of course, the same word-boundary-based representation of overlapping
turns could also be expressed in annotation graph form, by allowing
different speakers' transcripts to share certain nodes (representing
the word boundaries at which overlaps start or end). We do not suggest this,
since it seems to us to be based on an inappropriate model of overlapping,
which will surely cause trouble in the end.

Note the use of the \smtt{L/} `lexical' type to include the
full form of a contraction.  The UTF format employed special
syntax for expanding contractions.  No additional ontology
was needed in order to do this in the annotation graph.
Note also that it would have been possible to replicate the type
system, replacing \smtt{W/} with \smtt{W1/} for `speaker 1'
and \smtt{W2/} for `speaker 2'.  However, we have chosen instead
to attribute material to speakers using the \smtt{speaker/}
type on an arc spanning an entire turn.  The disconnectedness
of the graph structure means there can be no ambiguity about
the attribution of each component arc to a speaker.

As we have argued, annotation graphs of the kind shown in Figure~\ref{utf} are
actually more general and flexible than the UTF files they
model.  The UTF format imposes a linear sequence on the speaker
turns and complicates the transcript data of each turn with information
about overlap.
In contrast, the annotation graph structure provides a simple
representation for overlap, and it scales
up naturally to the situation where multiple speakers are
talking simultaneously, e.g.\ for transcribing a radio talk-back
show with a compere, a telephone interlocutor and a panel of discussants.

\subsection{Switchboard extensions}
\label{sec:switchboard}

The Switchboard corpus of conversational
speech \citep{Godfrey92} began with the three basic levels: conversation,
speaker turn, and word. Various parts of it have since been annotated
for syntactic structure \citep{Marcus93}, for breath groups and
disfluencies \citep{Taylor95}, for speech act type
\citep{JuretafskyBates97,JurafskyShriberg97}, and for phonetic
segments \citep{Greenberg96}. These various annotations have been done as separate
efforts, and presented in formats that are fairly easy to process one-by-one,
but difficult to compare or combine.  \citet{GraffBird00} provide
a detailed account of these multiple annotations of Switchboard.

\begin{figure}
{\tiny\setlength{\tabcolsep}{.25\tabcolsep}
\begin{tabular}{l|l|l}
\begin{minipage}[t]{.21\linewidth}
{\small\bf Aligned Word}
\begin{alltt}
B 19.44 0.16 Yeah,
B 19.60 0.10 no
B 19.70 0.10 one
B 19.80 0.24 seems
B 20.04 0.02 to
B 20.06 0.12 be
B 20.18 0.50 adopting
B 20.68 0.16 it.
B 21.86 0.26 Metric
B 22.12 0.26 system,
B 22.38 0.18 no
B 22.56 0.06 one's
B 22.86 0.32 very,
B 23.88 0.14 uh,
B 24.02 0.16 no
B 24.18 0.32 one
B 24.52 0.28 wants
B 24.80 0.06 it
B 24.86 0.12 at
B 24.98 0.22 all
B 25.66 0.22 seems
B 25.88 0.22 like.
A 28.44 0.28 Uh,
A 29.26 0.14 the,
A 29.48 0.14 the,
A 29.82 0.10 the
A 29.92 0.34 public
A 30.26 0.06 is
A 30.32 0.22 just
A 30.54 0.14 very
A 30.68 0.68 conservative
A 31.36 0.18 that
A 31.54 0.30 way
A 32.56 0.12 in
A 32.74 0.64 refusing
A 33.60 0.12 to
A 33.72 0.56 change
A 34.94 0.48 measurement
A 35.42 0.62 systems,
A 36.08 0.26 uh,
A 37.04 0.38 money,
A 37.62 0.30 dollar,
A 37.92 0.46 coins,
A 38.38 0.22 anything
A 38.60 0.18 like
A 38.78 0.30 that.
B 39.34 0.10 Yeah
B     *    * [laughter].
A 40.96 0.04 And,
A 41.32 0.04 and,
A 42.28 0.36 and
A 42.88 0.20 it
A     *    * [breathing],
A 43.08 0.16 it
A 43.48 0.46 obviously
A 43.94 0.22 makes
A 44.16 0.14 no
A 44.30 0.36 sense
A 44.66 0.06 that
A 44.72 0.12 we're
A 44.84 0.70 practically
A 46.52 0.32 alone
A 46.84 0.10 in
A 46.94 0.06 the
A 47.00 0.44 world
A 47.44 0.16 in,
A 48.52 0.04 in
A 48.56 0.26 using
A 48.82 0.08 the
A 48.90 0.22 old
A 49.12 0.40 system.
\end{alltt}
\end{minipage}
&
\begin{minipage}[t]{.2\linewidth}
{\small\bf Part of Speech}
\begin{alltt}
====================
[ SpeakerB22/SYM ]
./. 
====================

Yeah/UH ,/, 
[ no/DT one/NN ]
seems/VBZ to/TO
be/VB adopting/VBG 
[ it/PRP ] ./. 

[ Metric/JJ system/NN ]
,/, 
[ no/DT one/NN ]
's/BES very/RB ,/, 
[ uh/UH ] ,/, 
[ no/DT one/NN ]
wants/VBZ 
[ it/PRP ]
at/IN 
[ all/DT ]
seems/VBZ like/IN ./. 

====================
[ SpeakerA23/SYM ]
./. 
====================

[ Uh/UH ] ,/, 
[ the/DT ] ,/, 
[ the/DT ] ,/, 
[ the/DT public/NN ]
is/VBZ just/RB very/RB
conservative/JJ that/DT 
[ way/NN ]
in/IN refusing/VBG
to/TO change/VB 
[ measurement/NN
  systems/NNS ]
,/, 
[ uh/UH ] ,/, 
[ money/NN ] ,/, 
[ dollar/NN ] ,/, 
[ coins/NNS ] ,/, 
[ anything/NN ]
like/IN 
[ that/DT ] ./. 

====================
[ SpeakerB24/SYM ]
./. 
====================

Yeah/UH ./. 

====================
[ SpeakerA25/SYM ]
./. 
====================

And/CC ,/, and/CC ,/,
and/CC 
[ it/PRP ] ,/, 
[ it/PRP ]
obviously/RB makes/VBZ 
[ no/DT sense/NN ]
that/IN 
[ we/PRP ]
're/VBP practically/RB
alone/RB in/IN 
[ the/DT world/NN ]
in/IN ,/, in/IN
using/VBG 
[ the/DT old/JJ
  system/NN ]
./. 
\end{alltt}
\end{minipage}
&
\begin{minipage}[t]{.5\linewidth}
{\small\bf Disfluency}
\begin{alltt}
B.22:   Yeah, / no one seems to be adopting it. /
  Metric system, [ no one's very, + {F uh, } no one wants ]
  it at all seems like. / 
A.23:   {F Uh, } [ [ the, + the, ] + the ]
  public is just very conservative that way in
  refusing to change measurement systems,
  {F uh, } money, dollar, coins, anything like that. /
B.24:   Yeah <laughter>. /
A.25:   [ [ {C And, } +  {C and, } ] + {C and } ]
  [ it + <breathing>,  it ] obviously makes no sense
  that we're practically alone in the world [ in, + in ]
  using the old system. /
\end{alltt}

{\small\bf Treebank}
\begin{alltt}
((CODE SpeakerB22 .))
((INTJ Yeah , E_S))
((S (NP-SBJ-1 no one)
    (VP seems
        (S (NP-SBJ *-1) 
           (VP to (VP be (VP adopting (NP it)))))) . E_S))
((S (NP-TPC Metric system) ,
    (S-TPC-1 (EDITED (RM [)
                     (S (NP-SBJ no one)
                        (VP 's (ADJP-PRD-UNF very))) ,
                     (IP +)) (INTJ uh) ,
             (NP-SBJ no one)
             (VP wants (RS ]) (NP it) (ADVP at all)))
    (NP-SBJ *)
    (VP seems (SBAR like (S *T*-1))) . E_S))
((CODE SpeakerA23 .))
((S (INTJ Uh) ,
    (EDITED (RM [)
            (EDITED (RM [) (NP-SBJ-UNF the) , (IP +))
            (NP-SBJ-UNF the) , (RS ]) (IP +))
    (NP-SBJ-1 the (RS ]) public)
    (VP is
        (ADVP just)
        (ADJP-PRD very conservative)
        (NP-MNR that way)
        (PP in
            (S-NOM (NP-SBJ-2 *-1)
                   (VP refusing
                       (S (NP-SBJ *-2)
                          (VP to
                              (VP change
                                  (NP (NP measurement systems) ,
                                      (INTJ uh) , (NP money) ,
                                      (NP dollar) , (NP coins) ,
                                      (NP (NP anything)
                                          (PP like
                                              (NP that))))))))))) . E_S))
((CODE SpeakerB24 .))
((INTJ Yeah . E_S))
((CODE SpeakerA25 .))
((S (EDITED (RM [)
            (EDITED (RM [) And , (IP +)) and , (RS ]) (IP +)) and (RS ])
    (EDITED (RM [) (NP-SBJ it) (IP +) ,)
    (NP-SBJ (NP it)
            (SBAR *EXP*-1))
    (RS ])
    (ADVP obviously)
    (VP makes
        (NP no sense)
        (SBAR-1 that
                (S (NP-SBJ-2 we)
                   (VP 're
                       (ADVP practically) (ADJP-PRD alone)
                       (PP-LOC in (NP the world))
                       (EDITED (RM [) (PP-UNF in) , (IP +))
                       (PP in (RS ])
                           (S-NOM (NP-SBJ *-2)
                                  (VP using
                                      (NP the old system)))))))) . E_S))
\end{alltt}
\end{minipage}
\end{tabular}}
\vspace*{2ex}

\centerline{\epsfig{figure=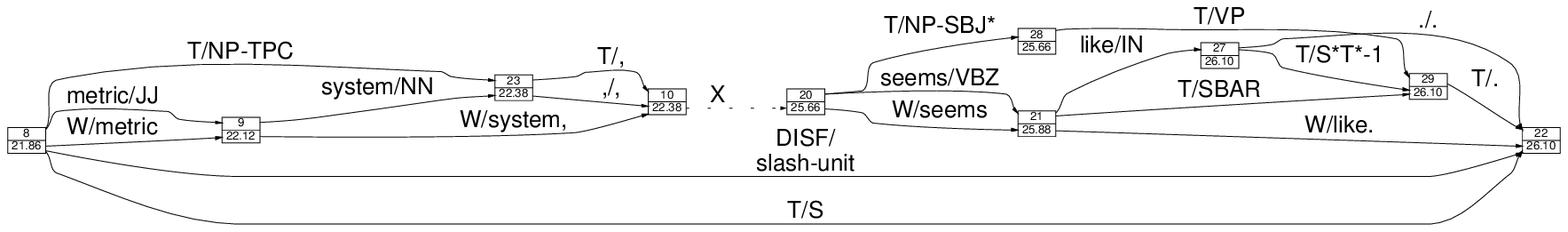,width=\linewidth}}
\vspace*{2ex}

\centerline{\epsfig{figure=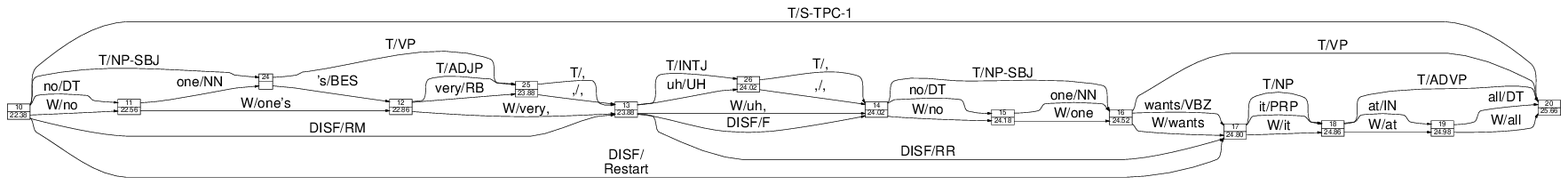,width=\linewidth}}
\caption{Multiple Annotations of the Switchboard Corpus, With Annotation Graph}\label{fig:swb}
\end{figure}

Figure~\ref{fig:swb} provides a
fragment of a Switchboard conversation, annotated for words,
part-of-speech, disfluency and syntactic structure.
Observe that punctuation is attached to the preceding
word in the case of word and disfluency annotation, while it
is treated as a separate element in the part-of-speech and Treebank
annotation.

Figure~\ref{fig:swb} also shows the annotation graph for this
Switchboard data, corresponding to
the interval [21.86, 26.10].  In this graph,
word arcs have type \smtt{W/},
Treebank arcs have \smtt{T/} and disfluency arcs
have \smtt{DISF/} type.  Types for the part-of-speech
arcs have been omitted for sake of clarity
(i.e. \smtt{Pos/metric/JJ} is written as just \smtt{metric/JJ}).
The graph is represented in two
pieces; the lower piece should be interpolated into the
upper piece at the position of the dotted arc labeled \smtt{X}.
Observe that the equivocation about
the status of punctuation is preserved in the annotation graph.

\subsection{MUC-7 Coreference Annotation}
\label{sec:coref}

\begin{figure}
\begin{sv}
<COREF ID="2" MIN="woman">This woman</COREF> receives three hundred dollars a month under
<COREF ID="5">General Relief</COREF>, plus <COREF ID="16" MIN="four hundred dollars"> four
hundred dollars a month in <COREF ID="17" MIN="benefits" REF="16">A.F.D.C.  benefits</COREF></COREF>
for <COREF ID="9" MIN="son"><COREF ID="3" REF="2">her</COREF>son</COREF>, who is
<COREF ID="10" MIN="citizen" REF="9">a U.S. citizen</COREF>.
<COREF ID="4" REF="2">She</COREF>'s among <COREF ID="18" MIN="aliens">an estimated five hundred
illegal aliens on <COREF ID="6" REF="5">General Relief</COREF> out of
<COREF ID="11" MIN="population"><COREF ID="13" MIN="state">the state</COREF>'s total illegal
immigrant population of <COREF ID="12" REF="11"> one hundred thousand </COREF></COREF></COREF>
<COREF ID="7" REF="5">General Relief</COREF> is for needy families and unemployable adults who
don't qualify for other public assistance.  Welfare Department spokeswoman Michael Reganburg says
<COREF ID="15" MIN="state" REF="13">the state</COREF> will save about one million dollars a year
if <COREF ID="20" MIN="aliens" REF="18">illegal aliens</COREF> are denied
<COREF ID="8" REF="5">General Relief</COREF>.
\end{sv}
\vspace*{2ex}

\centerline{\epsfig{figure=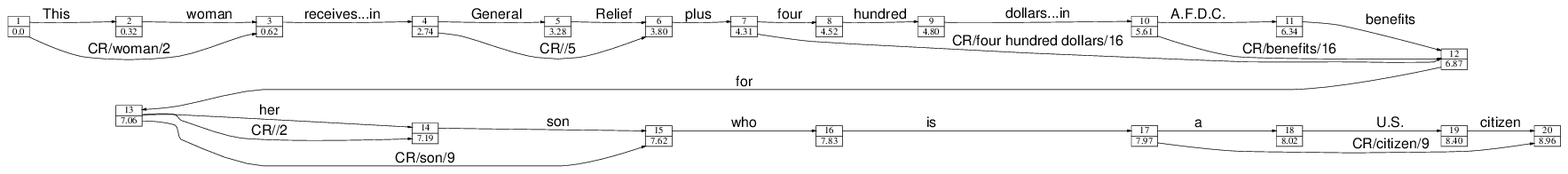,width=\linewidth}}
\caption{Annotation Graph for Coreference Example}\label{fig:coref}
\vspace*{2ex}\hrule
\end{figure}

The MUC-7 Message Understanding Conference specified
tasks for information extraction, named entity and coreference.
Coreferring expressions are to be linked
using SGML markup with \smtt{ID} and \smtt{REF} tags \citep{Hirschman97}.
Figure~\ref{fig:coref} is a sample of text from the
Boston University Radio Speech Corpus
[\myurl{www.ldc.upenn.edu/Catalog/LDC96S36.html}],
which has been marked up with coreference tags.

According to the MUC-7 specification,
noun phrases participating in coreference are wrapped with
\smtt{<coref>...</coref>} tags, and these can bear the attributes
\smtt{ID}, \smtt{REF}, \smtt{TYPE} and \smtt{MIN}.  
Each of these noun phrases is given a unique identifier, which may be
referenced by a \smtt{REF} attribute somewhere else.
Our example contains the following references:
$3 \rightarrow 2$,
$4 \rightarrow 2$,
$6 \rightarrow 5$,
$7 \rightarrow 5$,
$8 \rightarrow 5$,
$12 \rightarrow 11$,
$15 \rightarrow 13$,
$17 \rightarrow 16$.
The \smtt{TYPE} attribute encodes the relationship between the anaphor and
the antecedent.  Currently, only the identity relation is marked,
and so coreferences form an equivalence class.
Accordingly, our example contains the following equivalence classes:
$\left\{2, 3, 4\right\}$,
$\left\{5, 6, 7, 8\right\}$,
$\left\{11, 12\right\}$,
$\left\{13, 15\right\}$,
$\left\{16, 17\right\}$.
In our graph representation we have chosen the first number from each
of these sets as the identifier for the equivalence class, representing
it as the third attribute of an arc label.

\subsection{General Purpose Models}
\label{sec:general}

There are a number of existing annotation systems that are
sufficiently configurable that they can serve as general purpose
models for linguistic annotation.  Here we consider three such
systems: Emu, Festival and MATE.

The Emu speech database system \citep{CassidyHarrington00}
was designed to support speech scientists who work with large collections
of speech data, such as the
Australian National Database of Spoken Language
[\myurl{andosl.anu.edu.au/andosl/}].
Emu permits hierarchical annotations arrayed over any
number of levels, where each level is a linear ordering.
The levels and their relationships are fully customizable.

The Festival speech synthesis system
uses a data structure called a `heterogeneous relation
graph', which is a collection of binary relations
over feature structures (or attribute-value matrices)
\citep{Taylor00}.  Each feature structure
describes the local properties of some linguistic unit,
such as a segment, a syllable, or a syntactic phrase.
The value of an attribute could be atomic, or another feature
structure, or a function.
Functions have the ability to traverse one or more binary relations
and incorporate values from other feature structures.  A major use of these
functions is for propagating temporal information.

MATE is a dialogue annotation workbench based on XML and XSL
\citep{McKelvie00}.
Each layer of annotation is stored in a separate XML file, where
a layer could be a sequence of words or nested tags
representing a hierarchy.
Pieces of annotation reference each other using hyperlinks;
a tag can have a sequence of hyperlinks to represent a one-to-many
relationship.
MATE provides two ways to represent constituency --
nested tags (within a layer) and hyperlinks (between layers).
The structure of layers and their possible interrelationships is
highly configurable.

While these three models have important differences, all treat
the dominance relation as fundamental.  We believe this leads to
three problems of a non-trivial nature.

First, checking the temporal well-formedness
of an annotation requires navigating a potentially complex network
of multiple intersecting hierarchies.  In all three systems, this
checking task is simplified by storing the temporal information on
one level only (and possibly propagating the information outwards
from this level).  However this solution is inflexible with respect
to a common mode of corpus reuse, where an existing corpus with
temporal information on level $L_1$ is augmented with a new layer
$L_2$ of annotations which includes time offsets, and now the temporal
information must be coordinated across two (or more) levels.
In the general case of large, multi-layered annotations, it will
become computationally expensive to maintain temporal well-formedness,
even under very simple editing operations.  Perhaps for this reason,
none of the three models have been applied to large annotations.

A second problem concerns the representation of partial information.
As we shall see in \S\ref{sec:issues}, there are a variety of
situations where incomplete annotations arise, and where they should
be treated as well-formed despite only being partial.  However, both
Festival and MATE only permit complete well-formed hierarchies
to be represented and queried.  (Emu does not appear to have this
limitation.)

A third problem concerns expressive power.  In order to represent
intersecting hierarchies, these three systems employ pointer
structures (under the rubric of nested feature structures,
binary relations over feature structures,
XML nesting, hyperlinks, etc).  Yet this opens the door to
virtually any data structure, not just the kinds of annotations we saw
in \S\ref{sec:survey}.  This means that there are no general
properties of the model which can be exploited for efficient
computation.  Instead, users of these systems must keep the
annotations sufficiently small, or else the user interfaces must
ensure that the general purpose data structure is only used
in a restricted way.

We believe that it is preferable to adopt a
simpler model whose formal properties are well understood,
which is capable of representing multiple
hierarchies, and which foregrounds the temporal structure
of annotations.  Annotation graphs clearly meet these requirements.
They are sufficiently expressive to represent the diverse
range of annotation practice described in \S\ref{sec:survey},
and we believe their formal properties will facilitate the development of
scalable systems.

\section{Architectural Considerations}
\label{sec:arch}

A wide range of annotation models have now been considered, and
we have given a foretaste of the annotation graph model.
In this section we describe a
variety of architectural issues which we believe should be addressed
by any general purpose model for annotating linguistic signals.

\subsection{Various temporal and structural issues}
\label{sec:issues}

\subsubsection*{Partial Information}

In the discussion of CHILDES and the LACITO Archiving Project
above, there were cases where our graph representation
had nodes which bore no time reference.
Perhaps times were not measured, as in typical annotations
of extended recordings where time references might
only be given at major phrase boundaries (c.f.\ CHILDES).
Or perhaps time measurements were
not applicable in principle, as for phrasal translations
(c.f.\ the LACITO Archiving Project).
Various other possibilities suggest themselves.
We might create a segment-level annotation automatically from a word-level
annotation by looking up each word in a pronouncing dictionary
and adding an arc for each segment, prior to hand-checking
the segment annotations and adding time references to the
newly created nodes.  The annotation should remain well-formed
(and therefore usable) at each step in this enrichment process.

Just as the temporal information may be partial, so might
the label information.  For example, we might label indistinct
speech with whatever information is available -- `so-and-so said something
here that seems to be two syllables long and begins with a /t/'.

Beyond these two kinds of partiality, there is an even more
obvious kind of partiality we should recognize.
An annotated corpus might be annotated in a fragmentary
manner.  Perhaps only 1\% of a recording bears
on the research question at hand.  It should be possible
to have a well-formed annotation structure
with arbitrary amounts of annotation detail at certain
interesting loci, and limited or no detail elsewhere.
This is a typical situation in phonetic or sociolinguistic research,
where a large body of recordings may be annotated in detail with
respect to a single, relatively infrequent phenomenon of interest.

\subsubsection*{Redundant information}

An annotation framework (or its implementation) may also choose to
incorporate arbitrary amounts of redundant encoding of structural
information. It is often convenient to add redundant links explicitly
-- from children to parents, from parents to children, from one child
to the next in order, and so on -- so that a program can navigate the
structure in a way that is clearer or more efficient.  Although such
redundant links can be specified in the basic annotation itself
(cf.\ \citealt{Taylor00}) they might equally well be added
automatically, as part of a compilation or indexing process.  In our
view, the addition of this often-useful but predictable structure
should not be an intrinsic part of the definition of general-purpose
annotation structures.  We want to distinguish the annotation
formalism itself from various enriched data structures with redundant
encoding of hierarchical structure,
and from an application programming interface that
may dynamically compute and cache these enriched structures,
and from various indexes that support efficient access.

\subsubsection*{Multiple nodes at a time point}

\begin{figure}
\centerline{\epsfig{figure=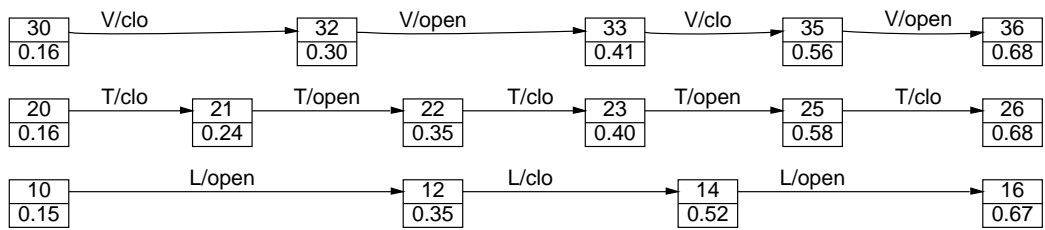,width=0.8\linewidth}}
\caption{Gestural Score for the Phrase 'ten pin'}\label{tenpin3}
\vspace*{2ex}\hrule
\end{figure}

In addition to hierarchical and sequential structure,
linguistic signals also exhibit parallel structure.  Consider the gestural
score notation used to describe the articulatory component
of words and phrases (e.g.\ \citealt{Browman89}).
A gestural score maps out the time course of the gestural
events created by the articulators of the vocal tract.
This representation expresses the fact that the articulators
move independently and that the segments we observe are the
result of particular timing relationships between the gestures.
Figure~\ref{tenpin3} gives an annotation graph for a gestural score.
The layers represent the velum \smtt{V/}, the tongue tip \smtt{T/} and the
lips \smtt{L/}.

Observe that nodes 12 and 22 have the same time reference.  This alignment
is a contingent fact about a particular utterance token.
An edit operation which changed the start time of one gesture would
usually carry no implication for the start time of some
other gesture.
Contrast this situation with a hierarchical structure, where, for example,
the left boundary of a
phrase lines up with the left boundary of its initial word.
Changing the time of the phrase boundary should change the
time of the word boundary, and vice versa.
In the general case, an update of this sort must propagate
both upwards and downwards in the hierarchy.
In fact, we argue that these two pieces of annotation actually
{\em share} the same boundary: their arcs emanate from a single
node.  Changing the time reference of that node does not
need to propagate anywhere, since the information is already
shared by the relevant arcs.

\subsubsection*{Instants}

Even though a linguistic event might have duration, such as the attainment
of a pitch target, the most perspicuous annotation may be tied
to an instant rather than an interval.  Some annotation
formalisms (e.g.\ Emu, Festival, Partitur)
provide a way to label instants.
The alignment of these instants with respect to other instants
or intervals can then be investigated or exploited.

We could extend our graph model to handle instants by
introducing labels on the nodes, or by allowing nodes to have
self-loops.  However, we prefer to give all label information
the same ontological status, and we are committed to the
acyclic graph model.  Therefore we adopt the following three
approaches to instants, to be selected as the situation dictates:
(i) instants can be treated as arcs between two nodes with
the same time reference; or
(ii) instants can be treated as short periods, where
these are labeled arcs just like any other; or
(iii) certain types of labels on periods could be interpreted
as referring to the commencement or the culmination of that period.
None of these require any extensions to the formalism.

\subsubsection*{Overlaps and gaps}

\begin{figure}[t]
\centerline{\epsfig{figure=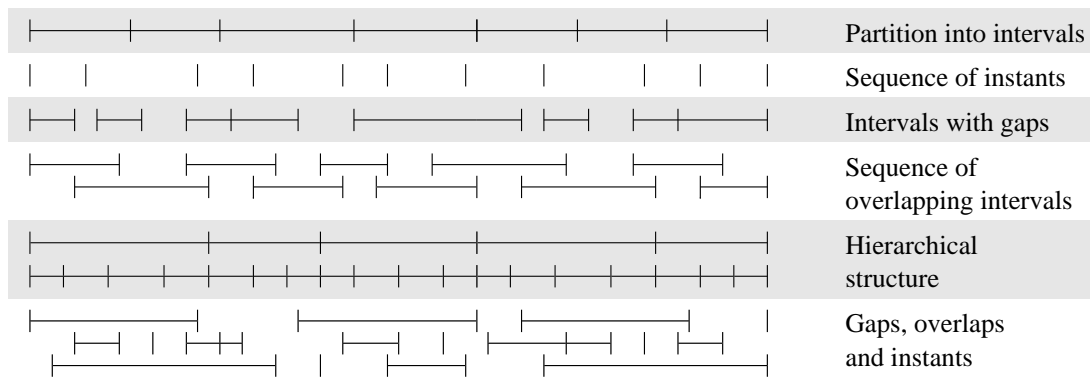,width=0.85\linewidth}}
\caption{Possible Structures for a Single Layer}\label{layer}
\vspace*{2ex}\hrule
\end{figure}

As we have seen, annotations are often stratified, where each
layer describes a different property of a signal.
What are the possible temporal relationships within
a given layer?  Some possibilities are
diagrammed in Figure~\ref{layer}, where a point is
represented as a vertical bar, and an interval is represented
as a horizontal line between two points.

In the first row of Figure~\ref{layer}, we see a layer which
exhaustively partitions the time-flow into a sequence
of non-overlapping intervals (or perhaps intervals which
overlap just at their endpoints).  In the second row we see
a layer of discrete instants.  The next two rows illustrate
the notions of gaps and overlaps.  Gaps might correspond to
periods of silence, or to periods in between the salient events,
or to periods which have yet to be annotated.
Overlaps occur between speaker turns in discourse (cf.\ Figure~\ref{callhome})
or even between adjacent words in a single speech stream
(cf.\ Figure~\ref{links}a).
The fifth row of Figure~\ref{layer}
illustrates a hierarchical grouping of intervals
within a layer.
The final row contains an arbitrary set of intervals and
instants.
We adopt this last option as the most general case for the layer of an annotation.
In other words, we impose no constraints on the structure of a layer.
In fact, layers themselves will not be treated specially;
a layer will be modeled as the collection of arcs having
the same type.

\subsection{Equivalence classes}
\label{sec:equivalence}

The arc data of an annotation graph is just a set.  Computationally, we can think
of it as an associative store -- just as in the
relational data model where ``tuples are identified through
a specification of their properties rather than by chasing pointers''
\cite[35]{Abiteboul95}.
There are cases where this structure appears inadequate,
and it seems necessary to enrich the ontology with inter-arc links.
This can be done by interpreting a particular field of an arc
label as a reference to some other arc.  However, in many cases,
including those discussed in this section, the links
are undirected (or the direction can be inferred) so we can treat them
as symmetric relations.  Transitivity seems harmless in these cases, and
so each mapping can be treated as an equivalence relation.
We consider three cases here, and the solution picks
up on the method which was used in \S\ref{sec:coref}.

Recall from Figure~\ref{tenpin3} that an annotation graph can
contain several independent streams of information,
where no nodes are shared between the streams.  The
temporal extents of the gestures in the different streams
are almost entirely asynchronous; any alignments are likely to be
coincidences.  However, these gestures may still have determinate, abstract
connections to elements of a phonological analysis. Thus a velar
opening and closing gesture may be associated with a particular nasal
feature, or with a set of nasal features, or with the sequence of
changes from non-nasal to nasal and back again.  But these associations
cannot usually be established purely as a matter of temporal
coincidence, since the phonological features involved are bundled
together into other units (segments or syllables or whatever)
containing other features that connect to other gestures whose
temporal extents are all different.  The rules
of coordination for such gestures involve phase
relations and physical spreading which are completely
arbitrary from the perspective of the representational framework.

\begin{figure}[t]
\centerline{\epsfig{figure=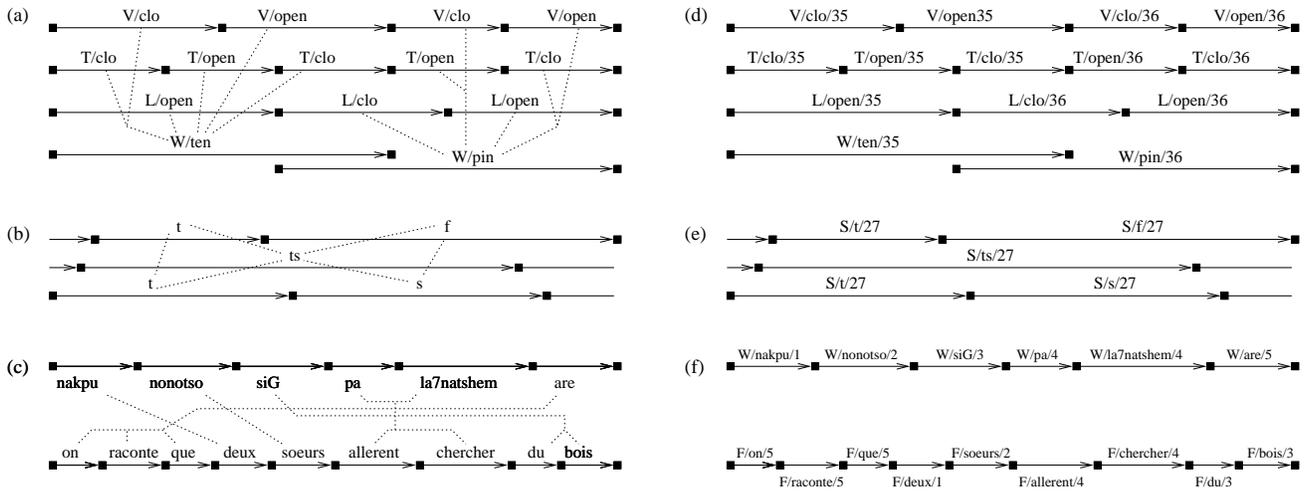,width=\linewidth}}
\caption{Inter-Arc Linkages Modeled Using Equivalence Classes}\label{links}
\vspace*{2ex}\hrule
\end{figure}

An example of the arbitrary relationship between
the gestures comprising a word is illustrated in
Figure~\ref{links}a.  We have the familiar annotation
structure (taken from Figure~\ref{tenpin3}), enriched with
information about which words license which gestures.
In the general case, the relationship between words
and their gestures is not predictable from the temporal structure
and the type structure alone.

The example in Figure~\ref{links}b shows a situation
where we have multiple independent
transcriptions of the same data.  In this case, the purpose
is to compare the performance of different transcribers
on identical material.
Although the intervals are not synchronized, it should be possible to navigate between
corresponding labels.

The final example, Figure~\ref{links}c, shows an annotation
graph based on the Hayu example from Figure~\ref{archivage}.
We would like to be able to represent the relationship between words of a phrasal
translation and the corresponding Hayu words.  This would be
useful, for example, for studying the various ways in which a particular
Hayu word is idiomatically translated.\footnote{
  The same linked multi-stream representation is
  employed in an actual machine translation system, \citealt{Brown90}.}
The temporal
relationship between linked elements is more chaotic here,
and there are examples of one-to-many and many-to-many
mappings.  In the general case, the words being mapped do not need
to be contiguous subsequences.

As stated above, we can treat all of these cases using equivalence classes.
Arcs are connected not by referencing one another,
but by jointly referencing a particular equivalence class.
For the gestural score in Figure~\ref{links}a, we assign each arc
to an equivalence class, as in Figure~\ref{links}d.  The class names
are arbitrary: in this case \smtt{35} and \smtt{36}.  Now we can easily
access the gestures licensed by a word regardless of their
temporal extent.  We can use type information to infer a
directionality for the association.  The same method works for the other cases,
and the proposed representations are shown in
Figure~\ref{links}e,f.
As a consequence of adopting this method,
there are now no less than three ways for a pair of arcs to be `associated':
temporal overlap, hierarchy, and a more abstract, atemporal relationship
(the equivalence-class linkages).
This three-way possibility mirrors the
three ways that ``autosegmental association'' is treated in the phonological literature
\citep{Bird95}.

\subsection{Hierarchical structure}
\label{sec:hierarchy}

Existing annotated speech corpora always involve a hierarchy of
several levels of annotation, even if they do not focus on very
elaborate types of linguistic structure.  TIMIT has sentences, words
and phonetic segments; a broadcast news corpus may have designated
levels for shows, stories, speaker turns, sentences and words. 
Some annotations may express much more elaborate hierarchies, with
multiple hierarchies sometimes created for a single underlying body of
speech data, such as Switchboard (see \S\ref{sec:switchboard}).

To represent hierarchical structure in the annotation graph model
we employ the parse chart construction
\cite[179ff]{GazdarMellish89}.  A parse chart is a particular kind of
acyclic digraph, which starts with a string of words and then adds a
set of arcs representing hypotheses about constituents dominating
various substrings.  Taking this as our starting point, we will
require that, for annotation graphs, if the substring spanned by arc
$a_i$ properly contains the substring spanned by arc $a_j$, then the
constituent corresponding to $a_i$ must dominate the constituent
corresponding to $a_j$ (though of course other structures may
intervene).  Hierarchical relationships are encoded
only to the extent that they are implied by this graph-wise inclusion
-- thus two arcs spanning the same substring are unspecified as to
their hierarchical relationship.
The graph structures implicit in TIMIT's annotation
files do not tell us, for the word spelled `I' and pronounced /ay/,
whether the word dominates the phoneme or vice versa; but the
structural relationship is implicit in the general relationship
between the two types of annotations.

We also need to mention that particular applications in
the areas of creation, query and display of annotations may
be most naturally organized in ways that motivate a user interface
based on a different sort
of data structure than the one we are proposing.
For instance, it may sometimes be easier to create
annotations in terms of tree-like dominance relations rather than chart-like
constituent extents, for instance in doing syntactic tree-banking
\citep{Marcus93}. It may likewise be easier in some cases to define 
queries explicitly in terms of tree structures. And finally, it may
sometimes be more helpful to display trees rather than equivalent
annotation graphs -- as done by some of the other general purpose
annotation models discussed in \S\ref{sec:general}.
We believe that such user interface issues will
vary from application to application, and may even depend on the
tastes of individuals in some cases. In any case, decisions about
such user interface issues are separable from decisions about
the appropriate choice of basic database structures.

\subsection{Discontinuous constituency}
\label{sec:discontinuous}

English lends itself to a description in terms of untangled
tree-structures, leaving a few phenomena (adverbials, parentheticals,
extraposed clauses, verb-associated particles, and so on) to be dealt
with in a way that violates canonical constituency.
In some languages, such as Latin, Czech, and Warlpiri,
it is common for several constituents to
be scrambled up together; the grammatical relations are
encoded using case marking.  Precisely for this reason, the surface syntax of
such languages seems to be best described in terms of
dependency relations, as opposed to constituent structures with no
constraints on string-tangling.  In the present context,
the point at issue is the following.
To what extent is it necessary for a treebanking
representation system to conveniently encode discontinuous
constituency?

\begin{figure}
\centerline{\epsfig{figure=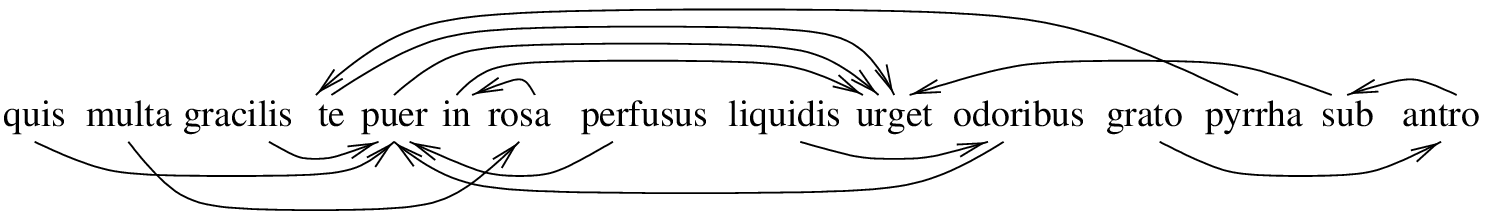,width=.8\linewidth}}
\vspace*{2ex}

\centerline{\epsfig{figure=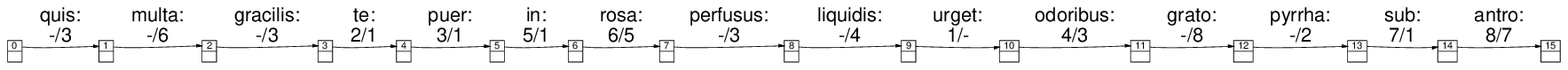,width=\linewidth}}

\vspace*{2ex}
\centerline{\epsfig{figure=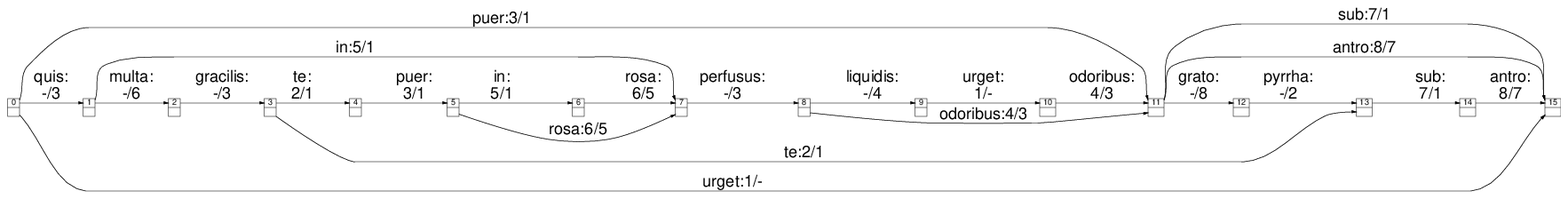,width=\linewidth}}

\caption{Sentence from Carmina 1.5 (Horace) Showing Dependency
Structure, with Two Annotation Graphs}
\label{fig:latin}
\vspace*{2ex}\hrule
\end{figure}

To date, few corpora have encoded discontinuous
constituency (see \citealt{Skut97} for an example), and so it would
be premature to propose a definitive answer to this question.
However, annotation graphs permit two representational possibilities,
both using the equivalence class construction.
The first possibility amounts to a version of dependency grammar,
while the second represents constituency in a manner that
reduces to the chart construction in cases where there are no
discontinuous constituents.
We illustrate the two possibilities
using a Latin sentence; see Figure~\ref{fig:latin}.

In the first (dependency grammar) version, each word arc carries
two additional fields.  The first field identifies the set of
dependents of the arc, while the second field identifies the
head of the arc.  In the second (constituency) version, the span
of a non-terminal is the smallest contiguous word string which
includes the words of its fringe.  In both cases, the numbers
are a direct representation of the constituency relation.

\subsection{Associations between annotations and files}
\label{sec:associations}

An `annotated corpus' is a set of annotation graphs and an associated
body of time series data.  The time series might comprise one or more
audio tracks, one or more video streams, one or more streams of
physiological data of various types, and so forth. The data might be
sampled at a fixed rate, or might consist of pairs of times and
values, for irregularly spaced times. Different streams will typically
have quite different sampling rates. Some streams might be defined
only intermittently, as in the case of a continuous audio recording
with intermittent physiological or imaging data. This is not an
imagined list of conceptually possible types of data -- we are familiar
with corpora with all of the properties cited.

It is not appropriate for an annotation framework to try to encompass
the syntax and semantics of all existing time series file
formats. They are simply too diverse and too far from being
stable. However, we do need to be able to specify what time series
data we are annotating, and how our annotations align with it, in a
way that is clear and flexible.

The time series data will be packaged into a set of one or more files.
Depending on the application, these files may have some more or less
complex internal structure, with headers or other associated
information about type, layout and provenance of the data. These
headers may correspond to some documented open standard, or they may
be embedded in a proprietary system.
The one thing that ties all of the time series data together is a
shared time base. To use these arbitrarily diverse data streams, we
need to be able to line them up time-wise. This shared time base is also
the only pervasive and systematic connection such data is likely to have
with annotations of the type we are discussing in this paper.
We will call this shared time base the ``timeline'', and ascribe it
formal status in the model.
Arbitrary additional information could be contained in the
internal structure of such time references, such as an
offset relative to the file's intrinsic time base (if any), or a
specification selecting certain dimensions of vector-valued data.

These timeline names will
permit an application to recover the time-series data that corresponds
to a given piece of annotation -- at least to the extent that the
annotation is time-marked and any time-function files have been
specified for the cited subgraph(s). Thus if time-marking is provided
at the speaker-turn level (as is often the case for published
conversational data), then a search for all the instances of a
specified word string will enable us to recover usable references to
all available time-series data for the turn that contains each of
these word strings. The information will be provided in the form of
timeline names, signal file names (and types where necessary),
time references, and perhaps time offsets; it will be the
responsibility of the application (or the user) to resolve these
references. If time-marking has been done at the word level, then the
same query will enable us to recover a more exact set of temporal
references into the same set of files.

The formalization of timelines is presented in \S\ref{sec:formalism}.
Our preference is to allow the remaining details of how to define
file references to fall outside the formalism.
It should be clear that there are simple and natural ways to
establish the sorts of linkages that are explicit in existing types of
annotated linguistic database.  After some practical experience,
it may make sense to try to provide a more formal account of references
to external time-series data.

\subsubsection*{Spatial and image-plane references}

We would also like to point out a wider problem for which we do not
have any general solution.  Although it is not our primary focus, we
would like the annotation formalism to be extensible to
spatially-specific annotations of video signals and similar data,
perhaps by enriching the
temporal anchors with spatial and/or image-plane information.  Anthropologists,
conversation analysts,
and sign-language researchers
are already producing annotations that are (at least
conceptually) anchored not only to time spans but also to a particular
spatial or image-plane trajectory through the corresponding series of video frames.

In the case of simple time-series annotations, we are tagging nodes with absolute
time references, perhaps offset by a single constant for
a given recorded signal.
However, if we are annotating a video recording,
the additional anchoring used for annotating video sequences
will mostly not be about absolute space, even with
some arbitrary shift of coordinate origin, but
rather will be coordinates in the image plane. If there
are multiple cameras, then image coordinates for each will differ,
in a way that time marks for multiple simultaneous recordings
do not.

In fact, there are some roughly similar cases in audio annotation,
where an annotation might reference some specific two- or
three-dimensional feature of (for instance) a time-series of
short-time amplitude spectra (i.e.\ a spectrogram), in which case the
quantitative details will depend on the analysis recipe. Our system
allows such references (like any other information) to be encoded in
arc labels, but does not provide any more specific support.

\subsubsection*{Relationship to multimedia standards}

In this context we ought to raise the question of how
annotation graphs relate to various multimedia standards like
the Synchronized Multimedia Integration Language
[\myurl{www.w3.org/TR/REC-smil/}]
and MPEG-4 [\myurl{drogo.cselt.it/mpeg/standards/mpeg-4/mpeg-4.htm}].
Since these provide ways to specify both
temporal and spatial relationships among strings, audio clips, still
pictures, video sequences, and so on, one hopes that they
will offer support for linguistic annotation. It is hard to offer a
confident evaluation, since MPEG-4 is still in development, and SMIL's
future as a standard is unclear. 

With respect to MPEG-4, we reserve judgment until its
characteristics become clearer.
Our preliminary assessment is that SMIL is not useful for purposes of
linguistic annotation, because it is mainly focused on presentational
issues (fonts, colors, screen locations, fades and animations, etc.)
and does not in fact offer any natural ways to encode the sorts of
annotations that we surveyed in the previous section. Thus it is easy
to specify that a certain audio file is to be played while a certain
caption fades in, moves across the screen, and fades out. It is not
(at least straightforwardly) possible to specify that a certain audio
file consists of a certain sequence of conversational turns,
temporally aligned in a certain way, which consist in turn of certain
sequences of words, etc.


\subsection{Node references versus byte offsets}
\label{sec:references}

The Tipster Architecture for linguistic annotation of text
is based on the concept of a fundamental, immutable textual foundation,
with all annotations expressed in terms of byte offsets into this
text \citep{Grishman96}. This is a reasonable solution for cases where the text
is a published given, not subject to revision by annotators.
However, it is not a good solution for speech transcriptions,
which are typically volatile entities, constantly up for revision both
by their original authors and by others.

In the case of speech transcriptions, it is more appropriate to treat
the basic orthographic transcription as just another annotation, no
more formally privileged than a discourse analysis or a translation.
Then we are in a much better position to deal with the common
practical situation, in which an initial orthographic transcription of
speech recordings is repeatedly corrected by independent users, who
may also go on to add new types of annotation of their own, and
sometimes also adopt new formatting conventions to suit their own
display needs. Those who wish to reconcile these independent
corrections, and also combine the independent additional annotations,
face a daunting task. In this case, having annotations reference byte
offsets into transcriptional texts is almost the worst imaginable
solution.

Although nothing will make it trivial to untangle this situation,
we believe our approach comes close.  As we shall see in \S\ref{sec:files},
our use of a flat, unordered file structure incorporating
node identifiers and time references means that
edits are as strictly local as they possibly can be, and
connections among various types of annotation are as durable as they
possibly can be. Some changes are almost completely transparent
(e.g.\ changing the spelling of a name). Many other changes will turn
out not to interact at all with other types of annotation. When there
is an interaction, it is usually the absolute minimum that is
necessary. Therefore, keeping track of what corresponds to what, across
generations of distributed annotation and revision, is as simple as
one can hope to make it.

Therefore we conclude that
Tipster-style byte offsets are an inappropriate choice for use as
references to audio transcriptions, except for cases
where such transcriptions are immutable in principle.

In the other direction, there are several ways to translate Tipster-style
annotations into our terms. The most direct way would be to treat Tipster byte
offsets exactly as analogous to time references -- since the only formal
requirement on our time references is that they can be ordered. This
method has the disadvantage that the underlying text could not be
searched or displayed in the same way that a speech transcription normally could.
A simple solution would be to add an arc for each of the lexical
tokens in the original text, retaining the byte offsets on the corresponding
nodes for translation back into Tipster-architecture terms.

\subsection{What is time?}
\label{sec:offsets}

TIMIT and some other extant databases denominate signal time in sample
numbers (relative to a designated signal file, with a known sampling
rate). Other databases use floating-point numbers, representing time
in seconds relative to some fixed offset, or other representations of
time such as centiseconds or milliseconds. In our formalization of
annotation graphs, the only thing that really matters about time
references is that they define an ordering. However,
for comparability across signal types, time references need to be
intertranslatable.

We feel that time in seconds is generally preferable to sample or
frame counts, simply because it is more general and easier to
translate across signal representations. However, there may be
circumstances in which exact identification of sample or frame numbers
is crucial, and some users may prefer to specify these directly to
avoid any possibility of confusion.  

Technically, sampled data points (such as audio samples or video
frames) may be said to denote time intervals rather than time points,
and the translation between counts and times may therefore become
ambiguous. For instance, suppose we have video data at 30 Hz. Should
we take the 30th video frame (counting from one) to cover the time
period from 29/30 to 1 second or from 29.5/30 to 30.5/30 second?  In
either case, how should the endpoints of the interval be assigned?
Different choices may shift the correspondence between times and frame
numbers slightly.

Also, when we have signals at very different sampling rates, a single
sampling interval in one signal can correspond to a long sequence of
intervals in another signal.
With video at 30 Hz and audio at 44.1 kHz, each video frame
corresponds to 1,470 audio samples. Suppose we have a time reference
of .9833 seconds.  A user might want to know whether this was
created because some event was flagged in the 29th video frame, for
which we take the mean time point to be 29.5/30 seconds, or because some
event was flagged at the 43,365th audio sample, for which we take the
central time point to be 43365.5/44100 seconds.

For reasons like these, some users might want the freedom to
specify references explicitly in terms of sample or frame numbers,
rather than relying on an implicit method of translation to and from
time in seconds.

\section{A Formal Framework}
\label{sec:algebra}

\subsection{Background}

All annotations of recorded linguistic signals require one unavoidable basic
action: to associate a label, or an ordered set of labels, with a
stretch of time in the recording(s). Such annotations also typically
distinguish labels of different types, such as spoken words vs.\ non-speech
noises. Different types of annotation often span different-sized
stretches of recorded time, without necessarily forming a strict
hierarchy: thus a conversation contains (perhaps overlapping)
conversational turns, turns contain (perhaps interrupted) words, and
words contain (perhaps shared) phonetic segments.

A minimal formalization of this basic set of practices is a directed
graph with fielded records on the arcs and optional time references on
the nodes.  We call these `annotation graphs' (AGs).
We believe that this minimal formalization in fact has
sufficient expressive capacity to encode, in a reasonably intuitive
way, all of the kinds of linguistic annotations in use today.  We also
believe that this minimal formalization has good properties with
respect to creation, maintenance and searching of annotations.

Our strategy is to see how far this simple conception can go,
resisting where possible the temptation to enrich its ontology
of formal devices, or to establish label types with special syntax or
semantics as part of the formalism.
It is important to recognize that translation into AGs
does not magically create compatibility among
systems whose semantics are different.  For instance, there are many
different approaches to transcribing filled pauses in English -- each
will translate easily into an AG framework, but their
semantic incompatibility is not thereby erased.

\subsection{Annotation graphs}
\label{sec:formalism}

\newtheorem{defn}{Definition}
\newtheorem{ex}{Example}

We take an annotation label to be a fielded record.
Depending on context, it is sometimes convenient to think of
such labels as an $n$-tuple of values distinguished by position, or
as a set of attribute-value pairs, or
as a set of functions from arcs to labels.
In this formalization we will adopt the first option, and employ
label sets $L_1, L_2, \ldots$, and adorn each arc with
a tuple of labels: $\left< l_1, l_2, \ldots \right>$.

The nodes $N$ of an AG reference signal data by virtue
of a function which maps nodes to time offsets.  An annotation may
reference more than one signal, and such signals may or may not
share the same abstract flow of time (e.g.\ two signals
originating from a stereo recording, versus two signals recorded
independently).  So we employ a collection of `timelines',
where each timeline is a totally ordered set.
AGs are now defined as follows:

\begin{defn}
An \textbf{annotation graph} $G$ over a label set $L$ and
timelines $\left<T_i, \leq_i\right>$ is a 3-tuple
$\left< N, A, \tau \right>$ consisting of a node set $N$,
a collection of arcs $A$ labeled with elements of $L$,
and a time function $\tau: N \rightharpoonup \bigcup T_i$,
which satisfies the following conditions:

\begin{enumerate}\setlength{\itemsep}{0pt}

\item $\left< N, A \right>$ is a labeled acyclic digraph
  containing no nodes of degree zero;

\item for any path from node $n_1$ to $n_2$ in $A$,
  if $\tau(n_1)$ and $\tau(n_2)$ are defined, then
  there is a timeline $i$ such that
  $\tau(n_1) \leq_i \tau(n_2)$.
\end{enumerate}
\end{defn}

Condition 1 requires that
each node of an AG is linked to at least one other node.
Note, however, that AGs may be disconnected (i.e.\ they may contain
disjoint sub-parts), and that they may be empty.
If $a = \left< n_1, l, n_2 \right>$ and $\tau(n_1) = \tau(n_2)$
then we call $a$ an instant.
It follows from the
second clause of this definition that any piece of {\it connected}
annotation structure can refer to at most one timeline.

Note that the interpretation of labels
as identifying substantive content, as conforming to a certain coding
standard, as meta-commentary on the annotation,
as signaling membership of some equivalence class, as referring to
material elsewhere (inside or outside the annotation), as an
anchor for an incoming cross-reference, as binary data, or
as anything else, falls outside the formalism.

We now illustrate this definition for the TIMIT graph in
Figure~\ref{timit}.  Let $L_1$ be the types of transcript
information (phoneme, word), and let
$L_2$ be the phonetic alphabet and the orthographic words used by TIMIT.
Let $T_1$ be the set of non-negative integers, the sample numbers.

{\small
\begin{eqnarray*}
N &=& \left\{ 0, 1, 2, 3, 4, 5, 6, 7, 8 \right\} \\
A &=&
\left\{
  \left< 0, \left< \mbox{P, h\#} \right>, 1 \right>,
  \left< 1, \left< \mbox{P, sh} \right>, 2 \right>,
  \left< 2, \left< \mbox{P, iy} \right>, 3 \right>,
  \left< 1, \left< \mbox{W, she} \right>, 3 \right>,
  \left< 3, \left< \mbox{P, hv} \right>, 4 \right>,
  \left< 4, \left< \mbox{P, ae} \right>, 5 \right>,
\right.\\&&\left.
  \left< 5, \left< \mbox{P, dcl} \right>, 6 \right>,
  \left< 3, \left< \mbox{W, had} \right>, 6 \right>,
  \left< 6, \left< \mbox{P, y} \right>, 7 \right>,
  \left< 7, \left< \mbox{P, axr} \right>, 8 \right>,
  \left< 6, \left< \mbox{W, your} \right>, 8 \right>
\right\} \\
\tau &=&
\left\{
  0 \rightarrow 0,     1 \rightarrow 2360,
  2 \rightarrow 3270,  3 \rightarrow 5200,
  4 \rightarrow 6160,  5 \rightarrow 8720,
  6 \rightarrow 9680,  7 \rightarrow 10173,
  8 \rightarrow 11077
\right\}
\end{eqnarray*}}

Next we define the notion of subgraphs.

\begin{defn}
An AG $\left< N', A', \tau' \right>$
is a \textbf{subgraph} of
an AG $\left< N, A, \tau \right>$ iff
$A' \subseteq A$; and
$N'$ and $\tau'$ are the restriction of $N$ and $\tau$
  to just those nodes used by $A'$.
If $G'$ is a subgraph of $G$ we write $G' \subseteq G$.
\end{defn}

Observe that the process of moving from an AG
to one of its subgraphs is fully determined by the selection of arcs.  There is no
freedom in the choice of the node set and the time function.
Therefore, we think of the subgraph relation as just
a {\it subset} relation on the arc set.

A {\it corpus} is just a set of
AGs along with a collection of signal files.
However, the division of a corpus into its
component annotations is somewhat arbitrary (cf.\ the division of
a text corpus into paragraphs, lines, words or characters).
For one operation we may want to view a speech corpus as a set of
speaker turns, where each turn is its own separate annotation
graph.  For a different operation it may be more natural to
treat the corpus as a set of broadcast programs, or a set of
words, or whatever.  Therefore we need to
blur the distinction between a single annotation and
a corpus of annotations.  But this is simple; the following
definition shows that a multi-annotation
corpus counts as a single annotation itself.

\begin{defn}
Let $G_1 = \left< N_1, A_1, \tau_1 \right>$
and $G_2 = \left< N_2, A_2, \tau_2 \right>$
be two AGs.  Then the \textbf{disjoint union}
of $G_1$ and $G_2$, written $G_1 \uplus G_2$,
is the AG
$\left< N_1 \uplus N_2, A_1 \cup A_2, \tau_1 \cup \tau_2 \right>$.
\end{defn}

So a corpus can be viewed either as a set of AGs, or as their
disjoint union.\footnote{
  Observe that the arc sets $A_i$ and the time functions
$\tau_i$ are guaranteed to be non-overlapping,
given that there can be no collision of elements of $N_1$ with $N_2$.
In practice, nodes will simply be assigned unique identifiers,
and these identifiers may be further qualified with a namespace.
In this way, while the internal structure of the corpus into individual
annotations might be reflected in file structure, it is
formally represented in the patterning of node identifiers.}

The result of a query against a
corpus is some subgraph of the disjoint union of the elements of
that corpus, which is itself an AG which can be treated as a derived corpus
and queried further.  Multiple independent
queries on the same corpus, or (equivalently) multiple corpora
derived from the same corpus, might then be combined by union, intersection
or relative complement.  The following definition is important for
the desired closure properties.
Let $2^G$ be the powerset of the AG $G$,
the set of subgraphs of $G$.

\begin{defn}\label{defn:algebra}
The \textbf{algebra} ${\cal A}_G$ of an AG
$G$ is the boolean algebra
\(
  \left<
    2^G,
    \cup,
    \cap,
    \bar{},
    \emptyset,
    G
  \right>
\),
where $\cup, \cap, \bar{}$ are set union, intersection
and (relative) complement, respectively.
Together with $\emptyset$ and $G$, these operations satisfy
the following identities:
$G_1 \cup \bar{G}_1 = G$,
$G_1 \cap \bar{G}_1 = \emptyset$, where $G_1 \subseteq G$.
Union and intersection also satisfy the usual distributive laws.
\end{defn}

Suppose we have a corpus containing a set of AGs $G_i$.
Let C = $\biguplus G_i$.
Then the space of all
possible query results for $C$ is $2^C$.
Now it is possible to endow a query language
with a model-theoretic semantics in terms of ${\cal A}_C$.

\subsection{Representation}
\label{sec:files}

Annotation graphs can be mapped to a variety of file formats,
including some of the formats described in our survey.  Here
we describe an XML `surface representation', which is maximally
flat and which makes explicit our intuition that AGs are
fundamentally a set of arcs.  Here we give an XML representation
for the above TIMIT example.\footnote{
  At the time of writing, a standard XML interchange format for annotation
  graphs is in development \citep{ATLAS00}.
}
The ordering of the arcs is not significant.

\begin{sv}
<annotation>
  <arc><source id="0" offset="0"/><label att_1="P" att_2="h\#"/><target id="1" offset="2360"/></arc>
  <arc><source id="1" offset="2360"/><label att_1="P" att_2="sh"/><target id="2" offset="3270"/></arc>
  <arc><source id="2" offset="3270"/><label att_1="P" att_2="iy"/><target id="3" offset="5200"/></arc>
  <arc><source id="1" offset="2360"/><label att_1="W" att_2="she"/><target id="3" offset="5200"/></arc>
  <arc><source id="3" offset="5200"/><label att_1="P" att_2="hv"/><target id="4" offset="6160"/></arc>
  <arc><source id="4" offset="6160"/><label att_1="P" att_2="ae"/><target id="5" offset="8720"/></arc>
  <arc><source id="5" offset="8720"/><label att_1="P" att_2="dcl"/><target id="6" offset="9680"/></arc>
  <arc><source id="3" offset="5200"/><label att_1="W" att_2="had"/><target id="6" offset="9680"/></arc>
  <arc><source id="6" offset="9680"/><label att_1="P" att_2="y"/><target id="7" offset="10173"/></arc>
  <arc><source id="7" offset="10173"/><label att_1="P" att_2="axr"/><target id="8" offset="11077"/></arc>
  <arc><source id="6" offset="9680"/><label att_1="W" att_2="your"/><target id="8" offset="11077"/></arc>
</annotation>
\end{sv}

In practice, the \smtt{id} and \smtt{offset}
attributes will be qualified with namespaces.
Offsets will be qualified with timeline information to identify a collection of
signal files sharing the same abstract timeline.  The ids will be qualified with
information about the annotation collection, sufficient to discriminate between
multiple independent annotations of the same signal data.  Under this scheme,
the name tag \smtt{<source id="5" time="8720"/>} might become:

\begin{sv}
<source id="http://www.ldc.upenn.edu/\~{}sb/timit-dr1-fjsp0#5"
        offset="TIMIT86://train/dr1/fjsp0#8720"/>
\end{sv}

The qualified node identifier now picks out the site, the annotator
\smtt{sb}, a
logical or physical name for the annotation, plus sufficient
information (here \smtt{\#5}) to pick out the node within that
annotation.\footnote{
  Note that frequently used namespaces can be defined once for
all as an XML entity and subsequently referenced using a much
shorter string (i.e. the entity reference).
}
Multiple annotations of the same signal data will
not overlap on these identifiers, and so they can be safely combined
into a single annotation if necessary.

The qualified time now identifies the corpus (a name which may need to
be resolved) and gives the path to the collection of signals sharing the
same timeline.  In the situation where multiple signals exist (as in
the case of multichannel recordings), the label data will specify the appropriate
signal(s).  Now multiple annotations of different signal data can
be safely combined into a single annotation if necessary.

As far as the annotation formalism is concerned, identifiers are
just unanalyzed strings.  Each timeline is a
separate $T_i$, and we simply have to guarantee that any pair of times
drawn from the same timeline can be compared using $\leq$.
(The comparison of times from separate timelines is not defined.)
The internal
syntax for identifiers and timelines is outside the formalism, as is
the rest of the above XML syntax (and any other syntax we may devise).
The main point here is that any reordering of arcs, any selection of
a subset of the arcs (via a query or some `grep'-like process), and
any concatenations of arc sets that came from the same corpus, are
well-formed as AG files.

\subsection{Anchored annotation graphs}

The nodes of an AG may or may not be {\it anchored} to a time point.
We now define an extension of AGs which constrains the
positions in which unanchored nodes can appear.

\begin{defn}
An \textbf{anchored annotation graph} is an AG where,
for any node $n$ that does not have both incoming and
outgoing arcs, then $\tau: n \mapsto t$ for some time $t$.
\end{defn}

Anchored AGs have no dangling arcs (or paths of arcs) leading
to an indeterminate time point.  It follows from this definition
that, for any unanchored node, we can reach an anchored node by
following a chain of arcs.  In fact {\it every} path from an unanchored
node will finally take us to an anchored node.  Likewise, an
unanchored node can be reached from an anchored node.
Thus, we are guaranteed to have temporal bounds for every node.
Observe that all AGs in \S\ref{sec:survey} are anchored.

Arbitrary subgraphs of anchored AGs may not be anchored, and so 
we cannot construct the algebra of an anchored AG.
In practice this is not a serious problem.  It is convenient for
annotated speech corpora to be anchored, since this greatly
facilitates speech playback and visual display.  The result
of querying an anchored AG will not generally be an anchored AG,
yet query results can be played back and graphically displayed in
the context of the original corpus, rather than in isolation.

Note that there is a special case where anchored AGs regain the
desired algebraic property:

\begin{defn}
A \textbf{totally-anchored AG} $G = \left< N, A, \tau \right>$ is an AG
where $\tau$ is total.
\end{defn}

In totally-anchored AGs, every node carries a time reference.
The AGs in Figures~\ref{timit} and \ref{partitur}
are all totally-anchored.

\subsection{Subsidiary relations on nodes and arcs}

As a further step towards the development of a query language, we can
define a variety of useful relations over nodes and arcs.

The first definition below allows us to talk about two kinds
of precedence relation on nodes in the graph structure.
The first kind respects the graph structure (ignoring the
time references), and is called
structural precedence, or simply {\it s-precedence}.  The second
kind respects the temporal structure (ignoring the graph
structure), and is called temporal precedence, or simply {\it t-precedence}.

\begin{defn}
A node $n_1$ \textbf{s-precedes} a node $n_2$,
written $n_1 <_s n_2$, if there is a path from $n_1$ to $n_2$.
A node $n_1$ \textbf{t-precedes} a node $n_2$,
written $n_1 <_t n_2$, if $\tau(n_1) < \tau(n_2)$.
\end{defn}

Observe that both these relations are transitive.
There is a more general notion of precedence which
mixes both relations.  For example, we can infer that
node $n_1$ precedes node $n_2$ if we can use a mixture
of structural and temporal information to get from $n_1$
to $n_2$.  This idea is formalized in the next definition.

\begin{defn}
\textbf{Precedence} is a binary relation on nodes, written $<$,
which is the transitive closure of the union of the s-precedes and the
t-precedes relations.
\end{defn}

This precedence relation is quadratic in the size
of the corpus, rendering it unusable in many situations.  However,
\citet{BirdBunemanTan00} have shown how this problem can be circumvented.

We can now define some useful
inclusion relations on arcs.
The first kind of inclusion respects the graph
structure, so it is called structural inclusion, or {\it s-inclusion}.
The second kind, {\it t-inclusion}, respects the temporal structure.
\begin{defn}
An arc $p = \left<n_1, n_4\right>$ \textbf{s-includes}
an arc $q = \left<n_2, n_3\right>$,
written $p \supset_s q$, if $n_1 <_s n_2$ and $n_3 <_s n_4$.
$p$ \textbf{t-includes} $q$, written $p \supset_t q$, if 
$n_1 <_t n_2$ and $n_3 <_t n_4$.
\end{defn}

As with node precedence, we define a general notion
of inclusion which generalizes over these two types:

\begin{defn}
\textbf{Inclusion} is a binary relation on arcs, written $\supset$,
which is the transitive closure of the union of the s-inclusion and the
t-inclusion relations.
\end{defn}

Note that all three inclusion relations are transitive.
We assume the existence of
non-strict precedence and inclusion relations, defined
in the obvious way.

The final definition concerns the
{\it greatest lower bound (glb)} and the {\it least upper bound (lub)}
of an arc.

\begin{defn}
Let $a = \left<n_1, l, n_2\right>$ be an arc.
\textbf{$\glb(a)$} is the greatest time value $t$ such that
there is some node $n$ with $\tau(n) = t$ and $n <_s n_1$.
\textbf{$\lub(a)$} is the least time value $t$ such that
there is some node $n$ with $\tau(n) = t$ and $n_2 <_s n$.
\end{defn}

According to this definition, the {\it glb} of an arc
is the time mark of the `greatest' anchored node from which the arc is
reachable.  Similarly, the {\it lub} of an arc is the time mark of
the `least' anchored node reachable from that arc.
The {\it glb} and {\it lub} are guaranteed to exist for
anchored annotation graphs, but not for annotation graphs in general.

\subsection{Multiple Annotations}

Linguistic analysis is always multivocal, in two senses. First, there
are many types of entities and relations, on many scales, from
acoustic features spanning a hundredth of a second to narrative
structures spanning tens of minutes. Second, there are many
alternative representations or construals of a given kind of
linguistic information.

Sometimes these alternatives are simply more or less convenient for a
certain purpose.  Thus a researcher who thinks theoretically of
phonological features organized into moras, syllables and feet, will
often find it convenient to use a phonemic string as a
representational approximation. In other cases, however, different
sorts of transcription or annotation reflect different theories about
the ontology of linguistic structure or the functional categories of
communication.

The AG representation offers a way to deal productively with both
kinds of multivocality. It provides a framework for relating different
categories of linguistic analysis, and at the same time to compare
different approaches to a given type of analysis. 

\begin{figure*}
\centerline{\epsfig{figure=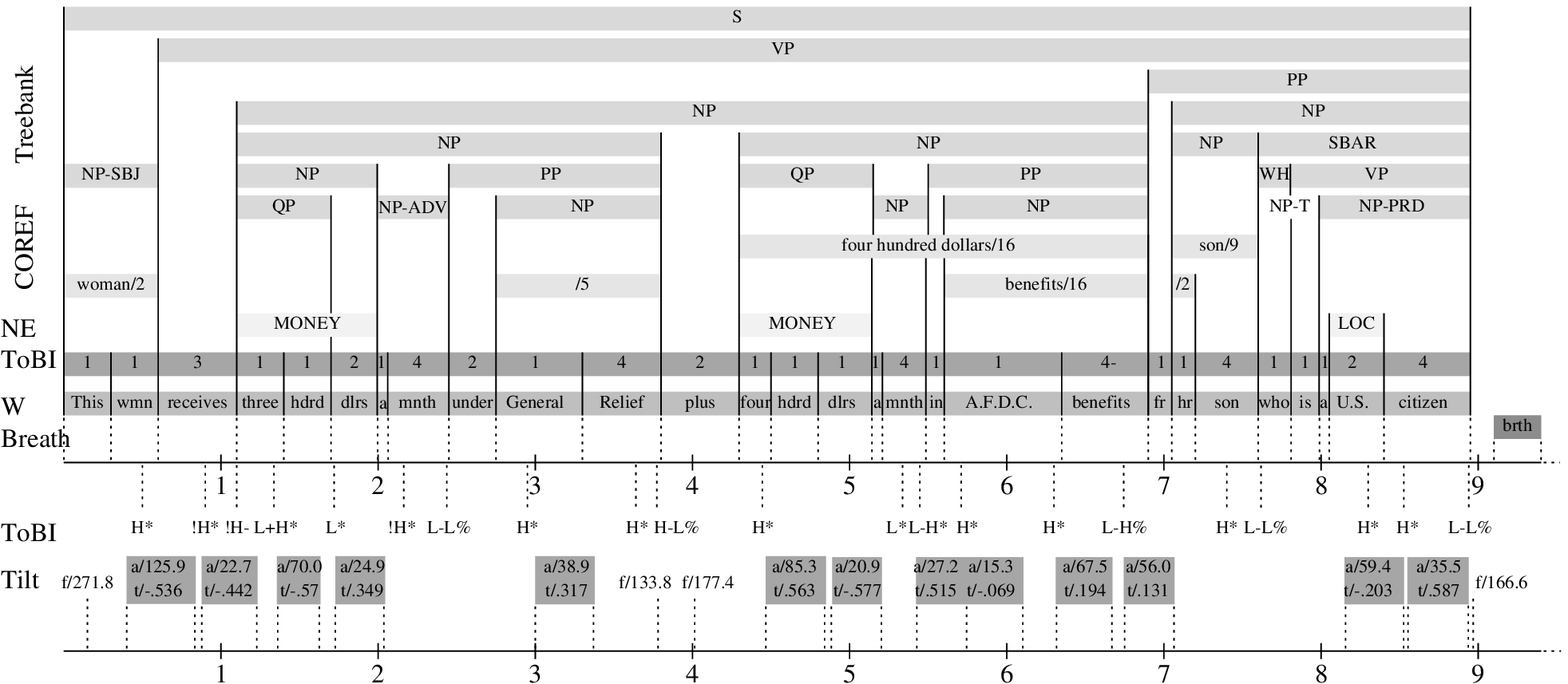,width=\linewidth}}
\caption{Visualization for BU Example}\label{fig:bu}
\vspace*{2ex}\hrule
\end{figure*}

As an example, Figure~\ref{fig:bu} shows a possible
AG-based visualization of eight
different sorts of annotation of a phrase from the BU Radio Corpus,
produced by Mari Ostendorf and others at Boston University, and
published by the LDC
[\myurl{www.ldc.upenn.edu/Catalog/LDC96S36.html}].
This multi-layer diagram corresponds to an annotation graph,
where arcs are represented by shaded rectangles, and
nodes are represented by solid vertical lines.
Anchored nodes are connected to a timeline with dotted lines,
and the point of intersection is labeled with a time reference.

The material in Figure~\ref{fig:bu} is from a recording of
a local public radio news broadcast. The BU annotations include four
types of information: orthographic transcripts, broad phonetic
transcripts (including main word stress), and two kinds of prosodic
annotation, all time-aligned to the digital audio files. The two kinds
of prosodic annotation implement the system known as ToBI
[\myurl{www.ling.ohio-state.edu/phonetics/E_ToBI/}].
ToBI is an acronym for ``Tones and Break Indices'', and correspondingly
provides two types of information: {\em Tones}, which are taken from a
fixed vocabulary of categories of (stress-linked) ``pitch accents'' and
(juncture-linked) ``boundary tones''; and {\em Break Indices}, which are
integers characterizing the strength and nature of interword
disjunctures.

We have added four additional annotations: coreference annotation and
named entity annotation in the style of MUC-7
[\myurl{www.muc.saic.com/proceedings/muc_7_toc.html}] provided
by Lynette Hirschman;
syntactic structures in the style of the Penn Treebank \citep{Marcus93}
provided by Ann Taylor;
and an alternative
annotation for the F$_0$ aspects of prosody, known as {\em Tilt}
\citep{Taylor98tilt} and
provided by its inventor, Paul Taylor. Taylor has done Tilt annotations
for much of the BU corpus, and intends to publish them as a point of
comparison with the ToBI tonal annotation. Tilt differs from ToBI in
providing a quantitative rather than qualitative characterization
of F$_0$ obtrusions: where ToBI might say ``this is a L+H* pitch accent,''
Tilt would say ``This is an F$_0$ obtrusion that starts at time $t_0$,
lasts for duration $d$ seconds, involves $a$ Hz total F$_0$ change,
and ends $l$ Hz different in F$_0$ from where it started.''

As usual, the various annotations come in a bewildering variety of
file formats. These are not entirely trivial to put into registration,
because (for instance) the Treebank terminal string contains both more
(e.g.\ traces) and fewer (e.g.\ breaths) tokens than the orthographic
transcription does. One other slightly tricky point: the connection
between the word string and the ``break indices'' (which are ToBI's
characterizations of the nature of interword disjuncture) are mediated
only by identity in the floating-point time values assigned to word
boundaries and to break indices in separate files. Since these time
values are expressed as ASCII strings, it is easy to lose the identity
relationship without meaning to, simply by reading in and writing out
the values to programs that may make different choices of internal
variable type (e.g.\ float vs.\ double), or number of decimal digits to
print out, etc.

Problems of this type are normal whenever multiple annotations need to
be compared. Solving them is not rocket science, but does take careful
work.  When annotations with separate histories involve mutually
inconsistent corrections, silent omissions of problematic material, or
other typical developments, the problems are multiplied.
In noting such difficulties, we are not criticizing the authors of the
annotations, but rather observing the value of being able to put
multiple annotations into a common framework.

Once this common framework is established, via translation of all eight
``strands'' into AG terms, we have the basis for posing queries
that cut across the different types of annotation.
For instance, we might look at the distribution of Tilt parameters
as a function of ToBI accent type; or the distribution of Tilt
and ToBI values for initial vs. non-initial members of coreference
sets; or the relative size of Tilt F0-change measures for nouns vs.
verbs.

\section{Conclusions and Future Work}
\label{sec:conclusion}

\subsection{Evaluation criteria}

There are many existing approaches to linguistic annotation, and many
options for future approaches.  Any evaluation of proposed frameworks,
including ours, depends on the costs and benefits incurred in a range
of expected applications. Our explorations have presupposed a
particular set of ideas about applications, and therefore a particular
set of goals.  We think that these ideas are widely shared, but it
seems useful to make them explicit.

Here we are using `framework' as a neutral term to encompass both the
definition of the logical structure of annotations, as discussed in this
paper, as well as various further specifications of e.g.\ annotation
conventions and file formats.

\begin{description}
\item[Generality, specificity, simplicity]\hfil\\

Annotations should be publishable (and will often be published), and
thus should be mutually intelligible across laboratories, disciplines,
computer systems, and the passage of time.

Therefore, an annotation framework should be sufficiently expressive
to encompass all commonly used kinds of linguistic annotation,
including sensible variants and extensions.  It should be capable of
managing a variety of (partial) information about labels, timing,
and hierarchy.

The framework should also be formally well-defined, and as simple as
possible, so that researchers can easily build special-purpose tools
for unforeseen applications as well as current ones, using future
technology as well as current technology.

\item[Searchability and browsability]\hfil\\

Automatic extraction of information from large annotation databases,
both for scientific research and for technological development, is a key
application.

Therefore, annotations should be conveniently and efficiently searchable,
regardless of their size and content. It should be possible to search
across annotations of different material produced by different groups
at different times -- if the content permits it -- without having to
write special programs. Partial annotations should
be searchable in the same way as complete ones.

This implies that there should be an efficient algebraic query
formalism, whereby complex queries can be composed out of well-defined
combinations of simple ones, and that the result of querying a set
of annotations should be just another set of annotations.

This also implies that (for simple queries) there should be
efficient indexing schemes, providing near constant-time access into
arbitrarily large annotation databases.

The framework should also support easy `projection' of natural
sub-parts or dimensions of annotations, both for searching and for
display purposes. Thus a user might want to browse a complex
multidimensional annotation database -- or the results of a
preliminary search on one -- as if it contained only an orthographic
transcription.

\item[Maintainability and durability]\hfil\\

Large-scale annotations are both expensive to produce and valuable to
retain. However, there are always errors to be fixed, and the
annotation process is in principle open-ended, as new properties can
be annotated, or old ones re-done according to new
principles. Experience suggests that maintenance of linguistic
annotations, especially across distributed edits and additions, can be
a vexing and expensive task.  Therefore, any framework should
facilitate maintenance of coherence in the face of distributed
correction and development of annotations.

Different dimensions of annotation should therefore be orthogonal, in
the sense that changes in one dimension (e.g.\ phonetic transcription)
do not entail any change in others (e.g.\ discourse transcription),
except insofar as the content necessarily overlaps.  Annotations of
temporally separated material should likewise be modular, so that
revisions to one section of an annotation do not entail global
modification.  Queries on material that is not affected by corrections or additions
should return the same thing before and after the updates.

In order to facilitate use in scientific discourse, it should
be possible to define durable references which remain valid wherever
possible, and produce the same results unless the referenced material
itself has changed.

Note that it is easy enough to define an invertible sequence of
editing operations for any way of representing linguistic annotations
-- e.g.\ by means of Unix `diff' -- but what we need in this case is
also a way to specify the correspondence (wherever it remains defined)
between arbitrary pieces of annotation before and after the
edit. Furthermore, we do not want to impose any additional burden on
human editors -- ideally, the work minimally needed to implement a
change should also provide any bookkeeping needed to maintain
correspondences.

\end{description}

How well does our proposal satisfy these criteria? 

We have tried to demonstrate generality, and to provide an adequate
formal foundation, which is also ontologically parsimonious (if not
positively miserly!).

Although we have not defined a query system, we have indicated the
basis on which one can be constructed: (tuple sets constituting)
AGs are closed under union, intersection and
relative complementation; the set of subgraphs of an
AG is simply the power set of its constituent tuples;
simple pattern matching on an AG can be
defined to produce a set of annotation subgraphs; etc.  Obvious sorts of
simple predicates on temporal relations, graphical relations, label
types, and label contents will clearly fit into this framework.

The foundation for maintainability is present: fully orthogonal
annotations (those involving different label types and time points) do
not interact at all, while linked annotations (such as those that
share time points) are linked only to the point that their content
requires. New layers of annotation can be added monotonically, without
any modification whatsoever in the representation of existing
layers. Corrections to existing annotations are as representationally
local as they can be, given their content.

Although we have not provided a recipe for durable citations (or for
maintenance of trees of invertible modifications), the properties just
cited will make it easier to develop practical approaches. In
particular, the relationship between any two stages in the development
or correction of an annotation will always be easy to compute as a set
of basic operations on the tuples
that express an AG. This makes it easy to calculate
just the aspects of a tree or graph of modifications that
are relevant to resolving a particular citation.

\subsection{Future work}
\label{sec:extensions}


\subsubsection*{Interactions with relational data}

Linguistic databases typically include important bodies of information
whose structure has nothing to do with the passage of time in any
particular recording, nor with the sequence of characters in
any particular text. For instance, the Switchboard corpus includes
tables of information about callers (including date of birth, dialect
area, educational level, and sex), conversations (including the
speakers involved, the date, and the assigned topic), and so on. This
side information is usually well expressed as a set of relational tables.
There also may be bodies of relevant information concerning a
language as a whole rather than any particular speech or text database:
lexicons and grammars of various sorts are the most obvious examples.
The relevant aspects of these kinds of information also
often find natural expression in relational terms.

Users will commonly want to frame queries that combine information of
these kinds with predicates defined on AGs: `find me all
the phrases flagged as questions produced by South Midland speakers
under the age of 30'.
The simplest way to permit this is simply to identify (some
of the) items in a relational database with (some of the) labels in an
annotation. This provides a limited, but useful, method for using the
results of certain relational queries in posing an annotational query,
or vice versa. More complex modes of interaction are also possible,
as are connections to other sorts of databases;
we regard this as a fruitful area for further research.

\subsubsection*{Generalizing time marks to an arbitrary ordering}

We have focused on the case of audio or video recordings, where a time
base is available as a natural way to anchor annotations. This role of
time can obviously be reassigned to any other well-ordered single
dimension.  The most obvious case is that of character- or
byte-offsets into an invariant text file. This is the principle used
in the Tipster Architecture \citep{Grishman96},
where all annotations
are associated with stretches of an underlying text, identified via
byte offsets into a fixed file. We do not think that this method is
normally appropriate for indexing into audio transcriptions, because they
are so often subject to revision (see \S\ref{sec:references}).

\subsubsection*{Generalizing node identifiers and arc labels}

As far as the formalism is concerned, the collection of node
identifiers and arc labels used in an AG are just sets. As a
practical matter, members of each set would obviously be represented
as strings. This opens the door to applications which encode
arbitrary information in these strings.
Indeed, the notion that arc labels encode `external' information is fundamental
to the whole enterprise.  After all, the point of the annotations is to
include strings interpreted as orthographic words, speaker names,
phonetic segments, file references, or whatever. These interpretations
are not built into the formalism, however, and this is an equally
important trait, since it determines the simplicity and generality of
the framework.

In the current formalization, arcs are decorated with fielded records.
This structure already contains a certain
amount of complexity, since the simplest kind of arc decoration would
be purely atomic.  In this case, we are convinced that the added value
provided by multiple fields is well worth the cost: all the bodies of
annotation practice that we surveyed had structure that was
naturally expressed in terms of atomic label types, and therefore a
framework in which arc decorations were just single uninterpreted
strings -- zeroth order labels -- would not be expressively adequate.
It is easy to imagine a wealth of other possible fields.
Such fields could identify the original annotator and the creation
date of the arc.
They could represent the confidence level of some other field.
They could encode a complete history of successive modifications.
They could provide
hyperlinks to supporting material (e.g.\ chapter and verse in the
annotators' manual for a difficult decision).
They could provide equivalence class identifiers (\S\ref{sec:equivalence}).
And they could include an arbitrarily-long SGML-structured commentary.

In principle, we could go still further, and decorate arcs with
arbitrarily nested feature structures endowed with a type
system \citep{Carpenter92} -- a second-order approach.  These
feature structures 
could contain references to other parts of the annotation, and
multiple structures could contain shared substructure.  These substructures
could be disjoined as well as conjoined, and appropriate features could
depend on the local type information.  A DTD-like label grammar could
specify available label types, their features and the type ordering.
We believe that this is a bad idea: it negates the effort that we made
to provide a simple formalism expressing the essential contents of
linguistic annotations in a natural and consistent way. Typed feature
structures are also very general and powerful devices, and entail
corresponding costs in algorithmic and implementational complexity.
Therefore, we wind up with a less useful representation that is much
harder to compute with.

Consider some of the effort that we have put into establishing a
simple and consistent ontology for annotation.  In the CHILDES case
(\S\ref{sec:childes}), we split a sentence-level annotation into a
string of word-level annotations for the sake of simplifying
word-level searches.  In the Switchboard Treebank case (\S\ref{sec:switchboard}) we
modeled hierarchical information using the syntactic chart
construction. Because of these choices, CHILDES and Switchboard
annotations become formally commensurate -- they can be searched or
displayed in exactly the same terms. With labels as typed feature
structures, a whole sentence, a complete tree structure, and indeed an entire
database could be packed into a single label. We could therefore have
chosen to translate CHILDES and Switchboard formats directly into typed
feature structures. If we had done this, however, the relationship
between simple concepts shared by the two formats -- such as lexical
tokens and time references -- would remain opaque.

Our preference is to extend the formalism cautiously, where it
seems that many applications will want a particular capability, and
to offer a simple mechanism to permit local or experimental extensions,
or approximations that stay within the confines of the existing formalism.

\subsection{Software}

We have claimed that AGs can provide an interlingua for
varied annotation databases, a formal foundation for queries on such
databases, and a route to easier development and maintenance of such
databases. Delivering on these promises will require software.
For those readers who agree with us
that this is an essential point, we will sketch our current perspective.

As our catalogue of examples indicated, it is fairly easy to translate
between other speech database formats and AGs, and we
have already built translators in several cases. We are also
developing software for creation, visualization,
editing, validation, indexing, and search, and have specified an
AG API and prototyped it in C++, with Perl and Tcl interfaces
\citep{ATLAS00}.
Our first goal is an open collection of relatively
simple tools that are easy to prototype and to modify, in preference
to a monolithic `annotation graph environment.'  However, we are
also committed to the idea that tools for creating and using
linguistic annotations should be widely accessible to computationally
unsophisticated users, which implies that eventually such tools need
to be encapsulated in reliable and simple interactive form.

Other researchers have also begun to experiment with the annotation
graph concept as a basis for their software tools, and a key index
of the idea's merit will of course be the extent to which tools are
provided by others.

\subsubsection*{Visualization, creation, editing}

Existing open-source software such as DGA Transcriber \citep{Barras98,Barras00},
and ISIP Transcriber [\myurl{www.isip.msstate.edu/resources/software/}],
whose user interfaces are all implemented in Tcl/tk,
make it easy to create interactive tools for creation, visualization,
and editing of AGs.
For instance, DGA Transcriber can be used without any changes to produce
transcriptions in the LDC Broadcast News format, which can then be translated
into AGs. Provision of simple input/output functions enables the
program to read and write AGs directly. The architecture
of the current tool is not capable of dealing with arbitrary
AGs, but a generalization of the software in that direction
is underway \citep{Geoffrois00}.

\subsubsection*{Validation}

An annotation may need to be submitted to a variety of validation
checks, for basic syntax, content and larger-scale structure.
First, we need to be able to tokenize and parse an annotation, without
having to write new tokenizers and parsers for each new task.
We also need to undertake some superficial syntax checking,
to make sure that brackets and quotes balance, and so on.
In the SGML realm, this need is partially met by DTDs.
We propose to meet the same need by developing
conversion and creation tools
that read and write well-formed graphs, and by input/output modules that
can be used in the further forms of validation cited below.

Second, various content checks need to be performed.  For instance,
are purported phonetic segment labels actually members of a designated
class of phonetic symbols or strings?  Are things marked as
`non-lexemic vocalizations' drawn from the officially approved list?
Do regular words appear in the spell-check dictionary?  Do capital letters
occur in legal positions?  These checks are not difficult to
implement,  e.g.\ as Perl scripts which use our AG API.

Finally, we need to check for correctness of hierarchies of arcs.
Are phonetic segments all inside words, which are all inside phrases,
which are all inside conversational turns, which are all inside conversations?
Again, it is easy to define such checks in a software environment
that has appropriately expressive primitives.

\subsubsection*{Indexing and Search}

A variety of indexing strategies for AGs would
permit efficient access to AG content centered on a temporal
locus, or based on the label information, or based on the hierarchies
implicit in the graph structure.  Such indexing is
well defined, algorithmically simple, and easy to implement in a
general way. Construction of general query systems, however,
is a matter that needs to be explored more fully in order to decide
on the details of the query primitives and the methods for building
complex queries, and also to try out different ways to express
queries. Among the many questions to be explored are:
how to express general graph- and time-relations;
how to integrate regular expression matching over labels;
how to integrate annotation-graph queries and relational queries;
how to integrate lexicons and other external resources; and
how to model sets of databases, each of which
contains sets of AGs, signals and perhaps relational
side-information.  Some of these issues are discussed further
by \citep{CassidyBird00,BirdBunemanTan00}.

It is easy to come up with answers to each of these questions,
and it is also easy to try the answers out, for instance in the context of
any system supporting the AG API.
We regard it as an open research problem to find good answers
that interact well, and also to find good ways to express queries in the
system that those answers will define.

\subsection{Envoi}

Whether or not our ideas are accepted by the various research
communities who create and use linguistic annotations, we hope to
foster discussion and cooperation among members of these communities.
A focal point of this effort is the Linguistic Annotation Page at
[\myurl{www.ldc.upenn.edu/annotation/}].

When we look at the numerous and diverse forms of linguistic
annotation documented on that page, we see underlying similarities
that have led us to imagine general methods for access and search, and
shared tools for creation and maintenance. We hope that this discussion
will move others in the same direction.

\section{Acknowledgements}

An earlier version of this paper was presented at ICSLP-98, and subsequently
as CIS Technical Report 99-01 [\myurl{xxx.lanl.gov/abs/cs.CL/9903003}].
Some material from a paper applying the model to discourse
\citep{BirdLiberman99dtag} was subsequently incorporated.
We are grateful to the following people for
discussions which have helped clarify our
ideas about annotations, and for comments on earlier drafts:
Claude Barras,
Sam Bayer,
Peter Buneman,
Jean Carletta,
Steve Cassidy,
Chris Cieri,
Hamish Cunningham,
David Day,
George Doddington,
John Garofolo,
Edouard Geoffrois,
David Graff,
Lynette Hirschman,
Ewan Klein,
Brian MacWhinney,
David McKelvie,
Boyd Michailovsky,
Michelle Minnick Fox,
Dick Oehrle,
Florian Schiel,
Richard Sproat,
Ann Taylor,
Paul Taylor,
Henry Thompson,
Marilyn Walker,
Peter Wittenburg,
Jonathan Wright,
Zhibiao Wu,
members of the Edinburgh Language Technology Group,
participants of the COCOSDA Workshop at ICSLP-98,
the Discourse Tagging Workshop at ACL-99,
and three anonymous reviewers.

\raggedright\small
\bibliographystyle{plainnat}
\bibliography{general}

\begin{thebibliography}{38}
\expandafter\ifx\csname natexlab\endcsname\relax\def\natexlab#1{#1}\fi

\bibitem[Abiteboul et~al.(1995)Abiteboul, Hull, and Vianu]{Abiteboul95}
Serge Abiteboul, Richard Hull, and Victor Vianu.
\newblock {\em Foundations of Databases}.
\newblock Addison Wesley, 1995.

\bibitem[Altosaar et~al.(1998)Altosaar, Karjalainen, Vainio, and
  Meister]{Altosaar98}
T.~Altosaar, M.~Karjalainen, M.~Vainio, and E.~Meister.
\newblock {Finnish} and {Estonian} speech applications developed on an
  object-oriented speech processing and database system.
\newblock In {\em Proceedings of the First International Conference on Language
  Resources and Evaluation Workshop: Speech Database Development for Central
  and Eastern European Languages}, 1998.
\newblock Granada, Spain, May 1998.

\bibitem[Anderson et~al.(1991)Anderson, Bader, Bard, Boyle, Doherty, Garrod,
  Isard, Kowtko, McAllister, Miller, Sotillo, Thompson, and
  Weinert]{Anderson91}
A.~Anderson, M.~Bader, E.~Bard, E.~Boyle, G.~M. Doherty, S.~Garrod, S.~Isard,
  J.~Kowtko, J.~McAllister, J.~Miller, C.~Sotillo, H.~Thompson, and R.~Weinert.
\newblock The {HCRC} {Map} {Task} corpus.
\newblock {\em Language and Speech}, 34:\penalty0 351--66, 1991.

\bibitem[Barras et~al.(1998)Barras, Geoffrois, Wu, and Liberman]{Barras98}
Claude Barras, Edouard Geoffrois, Zhibiao Wu, and Mark Liberman.
\newblock Transcriber: a free tool for segmenting, labelling and transcribing
  speech.
\newblock In {\em Proceedings of the First Language Resources and Evaluation
  Conference}, pages 1373--1376, 1998.

\bibitem[Barras et~al.(2000)Barras, Geoffrois, Wu, and Liberman]{Barras00}
Claude Barras, Edouard Geoffrois, Zhibiao Wu, and Mark Liberman.
\newblock Transcriber: development and use of a tool for assisting speech
  corpora production.
\newblock {\em Speech Communication}, 2000.
\newblock to appear.

\bibitem[Bird(1995)]{Bird95}
Steven Bird.
\newblock {\em Computational Phonology: A Constraint-Based Approach}.
\newblock Studies in Natural Language Processing. Cambridge University Press,
  1995.

\bibitem[Bird(1997)]{Bird97sigphon}
Steven Bird.
\newblock A lexical database tool for quantitative phonological research.
\newblock In {\em Proceedings of the Third Meeting of the ACL Special Interest
  Group in Computational Phonology}. Somerset, NJ: Association for
  Computational Linguistics, 1997.

\bibitem[Bird et~al.(2000{\natexlab{a}})Bird, Buneman, and
  Tan]{BirdBunemanTan00}
Steven Bird, Peter Buneman, and Wang-Chiew Tan.
\newblock Towards a query language for annotation graphs.
\newblock In {\em Proceedings of the Second International Conference on
  Language Resources and Evaluation}. Paris: European Language Resources
  Association, 2000{\natexlab{a}}.

\bibitem[Bird et~al.(2000{\natexlab{b}})Bird, Day, Garofolo, Henderson, Laprun,
  and Liberman]{ATLAS00}
Steven Bird, David Day, John Garofolo, John Henderson, Chris Laprun, and Mark
  Liberman.
\newblock Atlas: A flexible and extensible architecture for linguistic
  annotation.
\newblock In {\em Proceedings of the Second International Conference on
  Language Resources and Evaluation}. Paris: European Language Resources
  Association, 2000{\natexlab{b}}.

\bibitem[Bird and Liberman(1999)]{BirdLiberman99dtag}
Steven Bird and Mark Liberman.
\newblock Annotation graphs as a framework for multidimensional linguistic data
  analysis.
\newblock In {\em Towards Standards and Tools for Discourse Tagging --
  Proceedings of the Workshop}, pages 1--10. Somerset, NJ: Association for
  Computational Linguistics, 1999.
\newblock [xxx.lanl.gov/abs/cs.CL/9907003].

\bibitem[Browman and Goldstein(1989)]{Browman89}
Catherine Browman and Louis Goldstein.
\newblock Articulatory gestures as phonological units.
\newblock {\em Phonology}, 6:\penalty0 201--51, 1989.

\bibitem[Brown et~al.(1990)Brown, Cocke, {Della Pietra}, {Della Pietra},
  Jelinek, Mercer, and Roossin]{Brown90}
Peter~F. Brown, John Cocke, Stephen~A. {Della Pietra}, Vincent~J. {Della
  Pietra}, Fredrick Jelinek, Robert~L. Mercer, and Paul~S. Roossin.
\newblock A statistical approach to machine translation.
\newblock {\em Computational Linguistics}, 16:\penalty0 79--85, 1990.

\bibitem[Carpenter(1992)]{Carpenter92}
Bob Carpenter.
\newblock {\em The Logic of Typed Feature Structures}, volume~32 of {\em
  Cambridge Tracts in Theoretical Computer Science}.
\newblock Cambridge University Press, 1992.

\bibitem[Cassidy and Bird(2000)]{CassidyBird00}
Steve Cassidy and Steven Bird.
\newblock Querying databases of annotated speech.
\newblock In {\em Proceedings of the Eleventh Australasian Database
  Conference}, pages 12--20. Los Alamitos, CA: IEEE Computer Society, 2000.

\bibitem[Cassidy and Harrington(2000)]{CassidyHarrington00}
Steve Cassidy and Jonathan Harrington.
\newblock Multi-level annotation of speech: An overview of the emu speech
  database management system.
\newblock {\em Speech Communication}, 2000.
\newblock to appear.

\bibitem[Garofolo et~al.(1986)Garofolo, Lamel, Fisher, Fiscus, Pallett, and
  Dahlgren]{TIMIT86}
John~S. Garofolo, Lori~F. Lamel, William~M. Fisher, Jonathon~G. Fiscus,
  David~S. Pallett, and Nancy~L. Dahlgren.
\newblock {\em The {DARPA TIMIT} Acoustic-Phonetic Continuous Speech Corpus
  {CDROM}}.
\newblock NIST, 1986.
\newblock [www.ldc.upenn.edu/lol/docs/TIMIT.html].

\bibitem[Gazdar and Mellish(1989)]{GazdarMellish89}
Gerald Gazdar and Chris Mellish.
\newblock {\em Natural Language Processing in Prolog: An Introduction to
  Computational Linguistics}.
\newblock Addison-Wesley, 1989.

\bibitem[Geoffrois et~al.(2000)Geoffrois, Barras, Bird, and Wu]{Geoffrois00}
Edouard Geoffrois, Claude Barras, Steven Bird, and Zhibiao Wu.
\newblock Transcribing with annotation graphs.
\newblock In {\em Proceedings of the Second International Conference on
  Language Resources and Evaluation}. Paris: European Language Resources
  Association, 2000.

\bibitem[Godfrey et~al.(1992)Godfrey, Holliman, and McDaniel]{Godfrey92}
J.~J. Godfrey, E.~C. Holliman, and J.~McDaniel.
\newblock Switchboard: A telephone speech corpus for research and develpment.
\newblock In {\em Proceedings of the IEEE Conference on Acoustics, Speech and
  Signal Processing}, volume~I, pages 517--20, 1992.
\newblock [www.ldc.upenn.edu/Catalog/LDC93S7.html].

\bibitem[Graff and Bird(2000)]{GraffBird00}
David Graff and Steven Bird.
\newblock Many uses, many annotations for large speech corpora: Switchboard and
  tdt as case studies.
\newblock In {\em Proceedings of the Second International Conference on
  Language Resources and Evaluation}. Paris: European Language Resources
  Association, 2000.

\bibitem[Greenberg(1996)]{Greenberg96}
S.~Greenberg.
\newblock The switchboard transcription project.
\newblock LVCSR Summer Research Workshop, Johns Hopkins University, 1996.

\bibitem[Grishman(1997)]{Grishman96}
R.~Grishman.
\newblock {TIPSTER Architecture Design Document Version 2.3}.
\newblock Technical report, DARPA, 1997.
\newblock [www.nist.gov/itl/div894/894.02/related\_projects/tipster/].

\bibitem[Hertz(1990)]{Hertz90}
Susan~R. Hertz.
\newblock The delta programming language: an integrated approach to nonlinear
  phonology, phonetics, and speech synthesis.
\newblock In John Kingston and Mary~E. Beckman, editors, {\em Papers in
  Laboratory Phonology I: Between the Grammar and Physics of Speech},
  chapter~13, pages 215--57. Cambridge University Press, 1990.

\bibitem[Hirschman and Chinchor(1997)]{Hirschman97}
Lynette Hirschman and Nancy Chinchor.
\newblock {MUC}-7 coreference task definition.
\newblock In {\em Message Understanding Conference Proceedings}. Published
  online, 1997.
\newblock [www.muc.saic.com/proceedings/muc\_7\_toc.html].

\bibitem[Jacobson et~al.(2000)Jacobson, Michailovsky, and Lowe]{Jacobson00}
Michel Jacobson, Boyd Michailovsky, and John~B.\ Lowe.
\newblock Linguistic documents synchronizing sound and text.
\newblock {\em Speech Communication}, 2000.
\newblock to appear.

\bibitem[Jurafsky et~al.(1997{\natexlab{a}})Jurafsky, Bates, Coccaro, Martin,
  Meteer, Ries, Shriberg, Stolcke, Taylor, and {Van
  Ess-Dykema}]{JuretafskyBates97}
Daniel Jurafsky, Rebecca Bates, Noah Coccaro, Rachel Martin, Marie Meteer,
  Klaus Ries, Elizabeth Shriberg, Andreas Stolcke, Paul Taylor, and Carol {Van
  Ess-Dykema}.
\newblock Automatic detection of discourse structure for speech recognition and
  understanding.
\newblock In {\em Proceedings of the 1997 IEEE Workshop on Speech Recognition
  and Understanding}, pages 88--95, Santa Barbara, 1997{\natexlab{a}}.

\bibitem[Jurafsky et~al.(1997{\natexlab{b}})Jurafsky, Shriberg, and
  Biasca]{JurafskyShriberg97}
Daniel Jurafsky, Elizabeth Shriberg, and Debra Biasca.
\newblock {S}witchboard {SWBD-DAMSL} {L}abeling {P}roject {C}oder's {M}anual,
  {D}raft 13.
\newblock Technical Report 97-02, University of Colorado Institute of Cognitive
  Science, 1997{\natexlab{b}}.
\newblock [stripe.colorado.edu/\~{}jurafsky/manual.august1.html].

\bibitem[MacWhinney(1995)]{MacWhinney95}
Brian MacWhinney.
\newblock {\em The CHILDES Project: Tools for Analyzing Talk}.
\newblock Mahwah, NJ: Lawrence Erlbaum., second edition, 1995.
\newblock [childes.psy.cmu.edu/].

\bibitem[Marcus et~al.(1993)Marcus, Santorini, and Marcinkiewicz]{Marcus93}
Mitchell~P. Marcus, Beatrice Santorini, and Mary~Ann Marcinkiewicz.
\newblock Building a large annotated corpus of {English}: The {Penn}
  {Treebank}.
\newblock {\em Computational Linguistics}, 19\penalty0 (2):\penalty0 313--30,
  1993.
\newblock www.cis.upenn.edu/\~{}treebank/home.html.

\bibitem[McKelvie et~al.(2000)McKelvie, Isard, Mengel, Moller, Grosse, and
  Klein]{McKelvie00}
David McKelvie, Amy Isard, Andreas Mengel, M.~Moller, M.~Grosse, and Marion
  Klein.
\newblock The {MATE} workbench: an annotation tool for xml coded speech
  corpora.
\newblock {\em Speech Communication}, 2000.
\newblock to appear.

\bibitem[NIST(1998)]{UTF98}
NIST.
\newblock A universal transcription format {(UTF)} annotation specification for
  evaluation of spoken language technology corpora.
\newblock [www.nist.gov/speech/hub4\_98/utf-1.0-v2.ps], 1998.

\bibitem[Schegloff(1998)]{Schegloff98}
Emanuel Schegloff.
\newblock Reflections on studying prosody in talk-in-interaction.
\newblock {\em Language and Speech}, 41:\penalty0 235--60, 1998.
\newblock www.sscnet.ucla.edu/soc/faculty/schegloff/prosody/.

\bibitem[Schiel et~al.(1998)Schiel, Burger, Geumann, and Weilhammer]{Schiel98}
Florian Schiel, Susanne Burger, Anja Geumann, and Karl Weilhammer.
\newblock The {Partitur} format at {BAS}.
\newblock In {\em Proceedings of the First International Conference on Language
  Resources and Evaluation}, 1998.
\newblock [www.phonetik.uni-muenchen.de /Bas/BasFormatseng.html].

\bibitem[Skut et~al.(1997)Skut, Krenn, Brants, and Uszkoreit]{Skut97}
Wojciech Skut, Brigitte Krenn, Thorsten Brants, and Hans Uszkoreit.
\newblock An annotation scheme for free word order languages.
\newblock In {\em Proceedings of the Fifth Conference on Applied Natural
  Language Processing}, 1997.

\bibitem[Taylor(1995)]{Taylor95}
Ann Taylor.
\newblock {\em Dysfluency Annotation Stylebook for the Switchboard Corpus}.
\newblock University of Pennsylvania, Department of Computer and Information
  Science, 1995.
\newblock [ftp.cis.upenn.edu/pub/treebank/swbd/doc/DFL-book.ps], original
  version by Marie Meteer et al.

\bibitem[Taylor et~al.(2000)Taylor, Black, and Caley]{Taylor00}
Paul Taylor, Alan~W.\ Black, and Richard Caley.
\newblock Heterogeneous relation graphs as a formalism for representing
  linguistic information.
\newblock {\em Speech Communication}, 2000.
\newblock to appear.

\bibitem[Taylor(1998)]{Taylor98tilt}
Paul~A.\ Taylor.
\newblock The tilt intonation model.
\newblock In {\em Proceedings of the 5th International Conference on Spoken
  Language Processing}, 1998.

\bibitem[{Text Encoding Initiative}(1994)]{TEI-P3}
{Text Encoding Initiative}.
\newblock {\em Guidelines for Electronic Text Encoding and Interchange (TEI
  P3)}.
\newblock Oxford University Computing Services, 1994.
\newblock [www.uic.edu/orgs/tei/].

\end{thebibliography}

\end{document}